\documentclass[a4paper,10pt,twocolumn,preprint,3p]{elsarticle}
\pdfoutput=1   

\usepackage{subfig}   %
\usepackage{epstopdf} 
\usepackage{amssymb}
\usepackage{amsmath}
\usepackage{multicol}        
\usepackage{multirow}        
\usepackage[lined,boxed,linesnumbered]{algorithm2e}
\usepackage{float}
\usepackage{rotating}   
\usepackage[inline]{enumitem}
\usepackage{hyperref}
\usepackage{lineno}
\modulolinenumbers[50]

\makeatletter
\def\ps@pprintTitle{%
 \let\@oddhead\@empty
 \let\@evenhead\@empty
 \def\@oddfoot{}%
 \let\@evenfoot\@oddfoot}
\makeatother

\usepackage{numcompress}\biboptions{authoryear}

\hypersetup{pdfinfo={
Title={Visuospatial Skill Learning for Robots},
Author={S. Reza Ahmadzadeh},
Subject={Robotics},
Keywords={Robot Learning;Learning from Demonstration; imitation learning;visuospatial learning}
}}

\begin{document}

\begin{frontmatter}

\title{Visuospatial Skill Learning for Robots}

\author[rvt]{S. Reza Ahmadzadeh\corref{cor1}}
\ead{reza.ahmadzadeh@gatech.edu}
\author[focal]{Fulvio Mastrogiovanni}
\ead{fulvio.mastrogiovanni@unige.it}
\author[els]{Petar Kormushev}
\ead{p.kormushev@imperial.ac.uk}

\cortext[cor1]{Corresponding author}

\address[rvt]{School of Interactive Computing, Georgia Institute of Technology, Atlanta, GA, USA}
\address[focal]{Department  of  Informatics,  Bioengineering,  Robotics  and  Systems Engineering, University of Genova, via Opera Pia 13, 16145 Genova, Italy}
\address[els]{Dyson School of Design Engineering, Imperial College London, London SW7 2AZ, United Kingdom}

\begin{abstract}
A novel skill learning approach is proposed that allows a robot to acquire human-like visuospatial skills for object manipulation tasks. Visuospatial skills are attained by observing spatial relationships among objects through demonstrations. The proposed Visuospatial Skill Learning (VSL) is a goal-based approach that focuses on achieving a desired goal configuration of objects relative to one another while maintaining the sequence of operations. VSL is capable of learning and generalizing multi-operation skills from a single demonstration, while requiring minimum prior knowledge about the objects and the environment. In contrast to many existing approaches, VSL offers simplicity, efficiency and user-friendly human-robot interaction. We also show that VSL can be easily extended towards 3D object manipulation tasks, simply by employing point cloud processing techniques. In addition, a robot learning framework, VSL-SP, is proposed by integrating VSL, Imitation Learning, and a conventional planning method. In VSL-SP, the sequence of performed actions are learned using VSL, while the sensorimotor skills are learned using a conventional trajectory-based learning approach. such integration easily extends robot capabilities to novel situations, even by users without programming ability. In VSL-SP the internal planner of VSL is integrated with an existing action-level symbolic planner. Using the underlying constraints of the task and extracted symbolic predicates, identified by VSL, symbolic representation of the task is updated. Therefore the planner maintains a generalized representation of each skill as a reusable action, which can be used in planning and performed independently during the learning phase. The proposed approach is validated through several real-world experiments.
\end{abstract}

\begin{keyword}
robot learning\sep visuospatial skill\sep imitation learning\sep learning from demonstration\sep visual perception\sep
\end{keyword}

\end{frontmatter}

\nolinenumbers

\section{Introduction}
\label{sec:introduction}

During the past two decades, several robot skill learning approaches based on human demonstrations have been proposed \citep{ijspeert2013dynamical,argall2009survey}. Many of them address motor skill learning in which new sensorimotor skills are transferred to the robot using policy derivation techniques. Motor skill learning approaches can be categorized into two main groups: trajectory-based and goal-based. To emulate the demonstrated skill, the former group put the focus on recording and regenerating trajectories \citep{ijspeert2013dynamical} or intermittent forces \citep{kronander2012online}. For instance, \citet{bentivegna2004learning} proposed a method in which a humanoid robot learns to play air hockey by learning primitives such as velocity and positional trajectories. By combining Reinforcement Learning with Imitation Learning,~\citet{kormushev2010robot} taught a robot to flip pancakes, bootstrapping the learning process with a few observed demonstrations.

In many cases, however, it is not the trajectory that is of central importance, but the goal of the task (e.g. solving a jigsaw puzzle). Learning every single trajectory in such tasks actually, increases the complexity of the learning process unnecessarily \citep{niekum2012learning}. To address this drawback, the latter group of approaches which focus on the goal of the task has been proposed \citep{verma2005goal,dantam2012linguistic,chao2011towards}. There is a large body of literature on grammars from the Computational Linguistic and Computer Science communities, with a number of applications related to Robotics \citep{niekum2012learning,dantam2012linguistic}. Furthermore, a number of symbolic learning approaches exist, which focus on goal configuration rather than action execution \citep{chao2011towards}. However, in order to ground the symbols, they comprise many steps inherently, namely segmentation, clustering, object recognition, structure recognition, symbol generation, syntactic task modeling, motion grammar, and rule generation. Such approaches require a significant amount of \emph{a priori} knowledge to be manually engineered in the system \citep{niekum2012learning,dantam2012linguistic}. In addition, most above-mentioned approaches assume the availability of the information on the internal state of a tutor such as joint angles, which cannot be accessed easily by humans to imitate the observed behavior.

An alternative to motor skill learning is visual learning. These approaches are based on observing demonstrations and using human-like visual skills to replicate a task \citep{kuniyoshi1994learning,lopes2005visual}.

We propose a novel visual skill learning approach for interactive robot learning tasks. Unlike motor skill learning approaches, our framework utilizes visual perception as the main source of information for learning new skills from a demonstration. The proposed approach, which we refer to as Visuospatial Skill Learning (VSL), uses \emph{visuospatial skills} to replicate the demonstrated task. A \emph{visuospatial skill} is the capability to visually perceive the spatial relationship between objects. VSL requires minimum \emph{a priori} knowledge to learn a sequence of operations from a single demonstration. In contrast to many previous approaches, VSL exhibits simplicity, efficiency, and user-friendly human-robot interaction. Rather than relying on complicated models of human actions, labeled human data, or object recognition, our approach allows a robot to learn a variety of complex tasks effortlessly, simply by observing and reproducing the visual relationship among objects. We demonstrate the feasibility of the proposed approach in several experiments in which the robot learns to organize objects of different shape and color on a tabletop workspace to accomplish a goal configuration. In the conducted experiments the robot acquires and reproduces main capabilities such as absolute and relative object placement, classification, turn-taking, user intervention to modify the reproduction, and multiple operations performed on the same object.

In Section~\ref{sec:3dvsl}, we show that, by employing point cloud processing techniques instead of 2D image processing methods, VSL can be extended for object manipulation tasks in 3D space. The new variant of VSL is called VSL-3D.

Integrating VSL and a conventional motor learning approach such as Imitation Learning enables a robot to acquire new sensorimotor skills directly from demonstrations. In Section~\ref{sec:symbolic}, we show that by this integration, the need for a classical equation-based trajectory generation module can be avoided. In addition, in order to present the learned skills in a symbolic form, a standard action-level planner is combined with VSL. Learning preconditions and effects of the actions in VSL and making symbolic plans based on the learned knowledge deals with the long-standing important problem of Artificial Intelligence, namely Symbol Grounding~\citep{harnad1990symbol}. This framework is denoted as VSL-SP and is described in Section~\ref{sec:symbolic}.

In the rest of this paper, we use the term VSL for the original version of our proposed method, explained and used for 2D experiments in Section~\ref{sec:vsl}. We also use the term VSL-3D for the 3D extension of VSL, explained in Section~\ref{sec:3dvsl}. The proposed framework by integrating VSL, Imitation Learning, and symbolic planning explained in Section~\ref{sec:symbolic} is referred to as VSL-SP.

The paper is organized as follows. Related work is reviewed in Section~\ref{sec:relatedWork}. The terminology and methodology of the proposed approach together with a few implementation steps are explained in details in Section~\ref{sec:vsl}. Results for simulated and real-world experiments are reported in Section~\ref{subsec:simulation} and Section~\ref{subsec:realexp}, respectively. VSL-3D, the extension of the proposed approach for dealing with 3D applications, is presented in Section~\ref{sec:3dvsl}. Results reported in Section~\ref{subsec:experiment2} show that VSL-3D is readily attainable, although it requires more sophisticated image processing techniques. VSL-SP, the integrated framework, comprising a motor skill learning approach, VSL, and a symbolic planner, is described in Section~\ref{sec:symbolic}. The implementation steps for the proposed framework are discussed in Section~\ref{subsec:implementationSymbolic}. The feasibility and capability of the proposed framework are experimentally validated and the results are reported in Section~\ref{subsec:experiment3}. Conclusions of the paper are drawn in Section~\ref{sec:conclusions}.

\section{Related Work}
\label{sec:relatedWork}

Visual skill learning or \emph{learning by watching} is one of the most powerful mechanisms of learning in humans. It has been shown that even infants can imitate both facial and manual gestures~\citep{meltzoff1983newborn}. Learning by watching has been investigated in Cognitive Science as a source of higher order intelligence and fast acquisition of knowledge~\citep{rizzolatti1996premotor,park2008imitation}. In the next paragraph, related and most recent approaches on the topic of robot learning using visual perception are discussed.
Mentioned works include those that
\begin {enumerate*} [label=\itshape\upshape(\alph*\upshape)]
\item can be categorized as goal-based, \item utilize visual perception as the main source of information,  and \item especially those focusing on learning object manipulation tasks.
\end {enumerate*}

One of the most influential works on plan extraction has been discussed by~\citet{ikeuchi1994toward}. By obtaining and segmenting a continuous sequence of images their method extracts assembly plans. To extract transitions between sets of object relations, their approach depends on a model of the environment and predefined coordinate systems.

The early work of~\citet{kuniyoshi1994learning} focused on acquiring reusable high-level task knowledge by watching a demonstration. Their approach employs a model of the environment, a predefined action database, multiple vision sensors and basic visual features. Their approach cannot detect rotations and can only deal with rectangular objects which are not in contact with each other.

\citet{asada2000learning} proposed an approach based on \emph{epipolar constraint} for reconstructing the robot's view, on which adaptive visual servoing is applied to imitate an observed motion. Instead of using coordinate transformation, two sets of stereo cameras are used, one for the robot and the other for the demonstrator.

\citet{ehrenmann2001teaching} used multiple sensors for learning pick-and-place tasks in a kitchen environment. A magnetic field based tracking system, an active trinocular camera head, and a data glove were used in their experiments. Detected hand configurations were matched with a predefined symbol database detected using pre-trained neural networks.

\citet{yeasin2000toward} proposed a similar approach that focuses on extracting and classifying subtasks for grasping tasks. Their approach captures data by tracking the tutor's hand during demonstrations, generates trajectories, and extracts subtasks using neural networks. 

A visual learning by imitation approach presented by~\citet{lopes2005visual}, utilizes neural networks and viewpoint transformation to map visual perception to visuo-motor processes. They adopted a Bayesian formulation for gesture imitation. The method proposed by~\citet{pardowitz2006incremental} extracts knowledge about a pick-and-place task from a sequence of actions. They extract the proper primitive actions and the constraints of the task from demonstrations.

\citet{pastor2009learning} presented an approach to encoding motor skills into a non-linear differential equation that can reproduce those skills afterward. Object-Action Complex~\citep{kruger2009formal} was used to provide primitive actions with semantics. However, the resulting symbolic actions were never used in a high-level planner.

\citet{ekvall2008robot} proposed a symbolic learning approach in which a logical model for an STRIPS-like planner is learned from multiple human demonstrations. For each object, a geometric model and a set of features are given to the algorithm. The method can only deal with rectangular objects and the precision of the positioning is decided using the K-means clustering method. The spatial relations model presented by \citet{golland2010game} has several limitations which prevent it from being implemented on a real robot. For instance, the approach assumes a perfect and noiseless visual information and a perfect object segmentation. Furthermore, a small grammar is used, which means that the system just works for a few carefully constructed expressions.

\citet{chao2011towards} proposed a symbolic goal-based approach for grounding discrete concepts from continuous perceptual data using unsupervised learning. Task goals are learned using Bayesian inference. The approach was implemented on tasks including a single pick-and-place operation in which the initial objects configuration is randomized.

\citet{mason2011robot} used verbal commands to instruct a robot to perform a room tidying task. Primitive actions are predefined and the approach uses a database including all possible discrete configurations of objects. The features are learned using a beta regression classifier and the planner finds a solution using simulated annealing. The conducted experiments are of particular value for us because they can be easily learned and reproduced using our approach.

\citet{kroemer2011flexible} proposed a two-layer framework for learning manipulation tasks. One layer optimizes the parameters of motor primitives, whereas the other relies on pre- and post-conditions to model actions. In contrast to VSL, pre- and post-conditions are manually engineered in the system. \citet{aksoy2011learning} showed how symbolic representations can be linked to the trajectory level using spatiotemporal relations between objects and hands in the workspace. In a similar way, \citet{aein2013toward} showed how such grounded symbolic representations can be employed together with low-level trajectory information to create action libraries for imitating observed actions.

There is a large body of literature on grammars from the Computational Linguistic and Computer Science communities, with a number of applications related to Robotics~\citep{niekum2012learning,dantam2012linguistic}. ~\citet{niekum2012learning} proposed a goal-based method for learning from trajectory-based demonstrations. Segmentation and recognition are achieved using a Beta-Process Autoregressive Hidden Markov Model (HMM), while Dynamic Movement Primitives are used to reproduce the skill. The approach needs to assign a coordinate frame to each detected object and it cannot deal with rotations. Increasing the number of operations in the demonstrations makes the segmentation process increasingly more prone to errors. Sometimes during the segmentation process, extra skills can be extracted mistakenly.

\citet{dantam2011chess} proposed Motion Grammar to plan the operation of a robot through a context-free grammar. The method was used to learn human-robot chess games in which a point cloud of the chessboard was used to detect the pieces.
\citet{dantam2012linguistic} presented an approach for the visual analysis of demonstrations and automatic policy extraction for an assembly task. They used several techniques such as segmentation, clustering, symbol generation and abstraction. \citet{guadarrama2013grounding} proposed a natural language interface for grounding nouns and spatial relations. The system uses a classifier trained using a huge database of collected images for each object. The presented applications and the spatial representations are limited to 2D information.

By introducing a stack-based domain specific language for describing object repositioning tasks, \citet{feniello2014program} built a framework based on our initial VSL approach~\citep{ahmadzadeh2013visuospatial}. By performing demonstrations on a tablet interface, the time required for teaching is reduced and the reproduction phase can be validated before execution of the task in the real-world. Various types of real-world experiments have been conducted including sorting, kitting, and packaging tasks.

There is a long history of research in the area of interpreting spatial relations~\citep{burgard1999experiences,kelleher2006proximity,moratz2006spatial,skubic2004spatial}. In most of these works, a model of spatial relations is built by hard-coding the meaning of the spatial relations. Instead, some recent works concluded that a learned model of spatial relations can outperform hard-coded models~\citep{golland2010game,guadarrama2013grounding}.

Visuospatial Skill Learning (VSL), is a goal-based approach that utilizes visual perception as the main source of information. It focuses on achieving the desired goal configuration of objects relative to one another while maintaining the sequence of operations. We show that VSL is capable of learning and generalizing multi-operation skills from a single demonstration while requiring minimum prior knowledge about the objects and the environment. We also show that VSL can be extended towards 3D object manipulation tasks, simply by employing point cloud processing techniques. In contrast to many existing approaches, VSL offers simplicity, efficiency and user-friendly human-robot interaction. In addition, we propose a learning framework (VSL-SP) by integrating, imitation learning, VSL, and a conventional planning method. By integrating VSL with a conventional trajectory-based learning approach, we show that primitive actions can be learned directly through demonstrations instead of being programmed manually in the system. The proposed integrated approach easily extends and adapts robot capabilities to novel situations, even by users without programming ability. To utilize the advantages of standard symbolic planners, in VSL-SP the internal planner is integrated with an existing action-level symbolic planner. The planner represents a symbolic description of the skills. It uses the underlying constraints of the task and extracted symbolic predicates (i.e. action preconditions and effects), identified by VSL, it updates the symbolic representation while the skills are being learned. Therefore the planner maintains a generalized representation of each skill as a reusable action, which can be used in planning and performed independently during the learning phase. We validate our approach through several real-world experiments with a Barret WAM manipulator and an iCub humanoid robot.

\section{Introduction to Visuospatial Skill Learning}
\label{sec:vsl}

\begin{figure}[ht]
    \subfloat[Virtual experimental setup for a VSL task consisting of a robot manipulator, a vision sensor, a set of objects, and a workspace.]
    {
        \includegraphics[width=0.9\columnwidth]{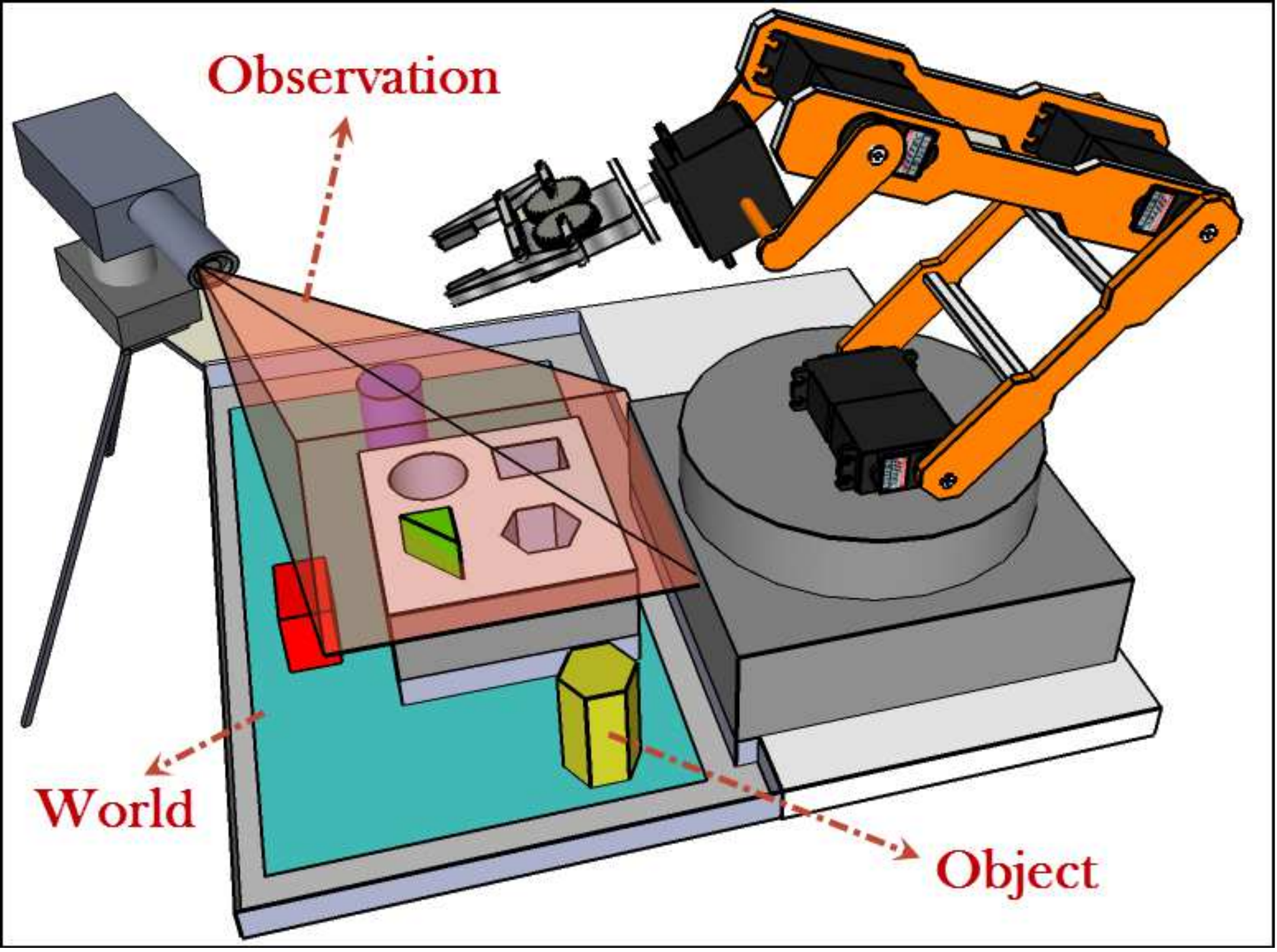}
        \label{fig:virtualarm}
    }
    \\
    \subfloat[A high-level flow diagram of VSL illustrating the demonstration and reproduction phases.]
    {
        \includegraphics[width=0.9\columnwidth]{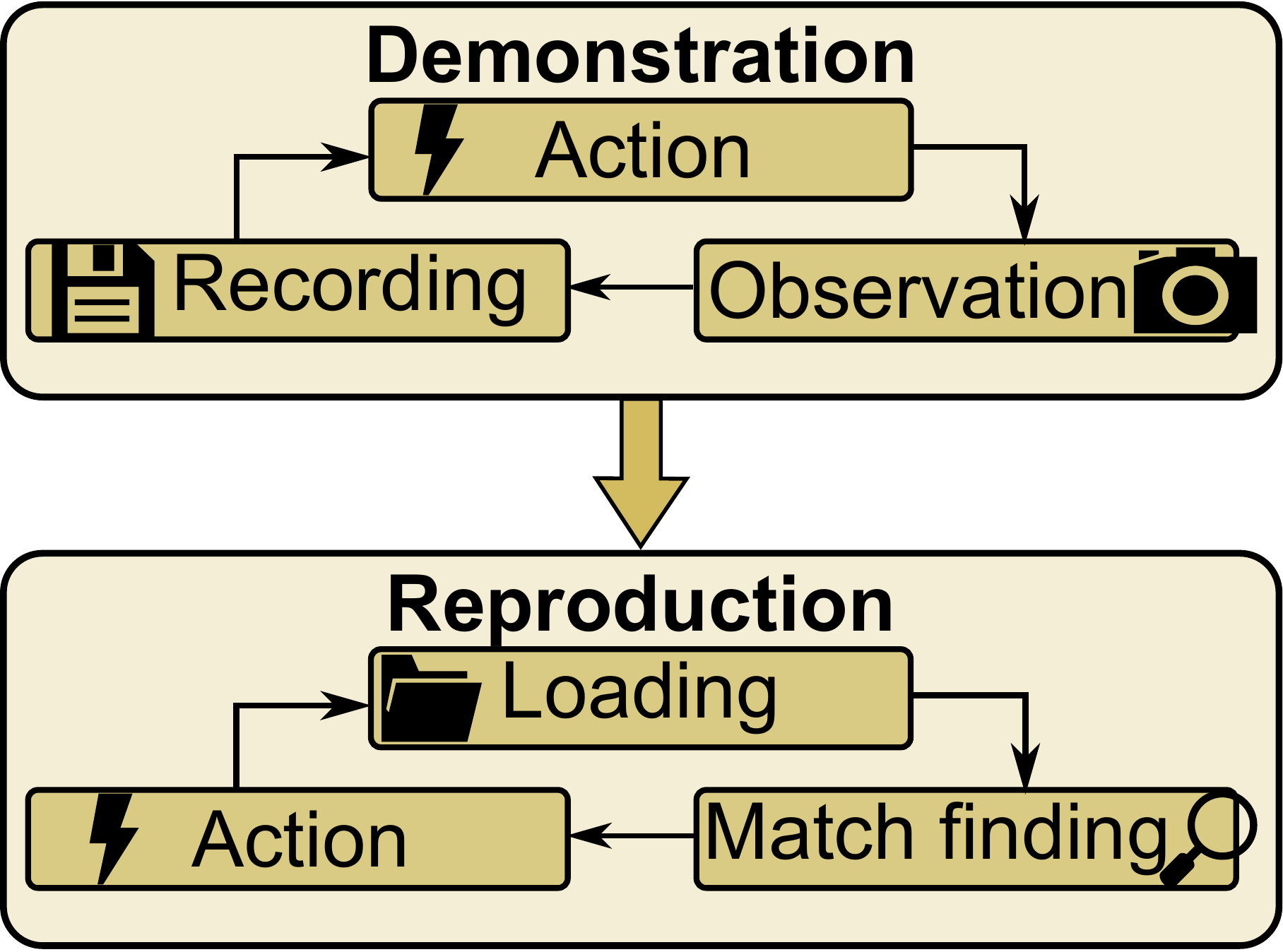}
        \label{fig:mainvsl}
    }
    \caption{Virtual setup and flow diagram for VSL.}
    \label{fig:vsetupflow}
\end{figure}

Visuospatial Skill Learning (VSL) is a goal-based robot Learning from Demonstration approach. A tutor demonstrates a sequence of operations on a set of objects. Each operation consists of a set of actions (e.g. pick and place). In this Section, we consider pick-and-place object manipulation tasks in which achieving the goal of the task and retaining the sequence of operations are particularly important. In Section~\ref{sec:symbolic}, we also consider other actions such as push and pull. We assume the virtual experimental setup illustrated in Figure~\ref{fig:virtualarm}, which consists of a robot manipulator equipped with a gripper, a tabletop workspace, a set of objects, and a vision sensor. The sensor can be mounted above the workspace to observe the tutor performing the task. Using VSL, the robot learns new skills from demonstrations by extracting spatial relationships among objects. Afterward, starting from a random initial objects configuration, the robot can perform a new sequence of operations resulting in reaching the same goal as the one demonstrated by the tutor. A high-level flow diagram, shown in Figure~\ref{fig:mainvsl}, illustrates the two main phases of VSL, namely demonstration and reproduction. In the demonstration phase, for each action, a set of observations is recorded used for the match finding process during the reproduction phase. During the match finding process, observations including the similar set of features are matched. In this Section, first, the basic terms for describing VSL are defined. Then, the problem statement is described and finally, the VSL approach is explained.

\subsection{Terminology}
\label{subsec:terminology}

In order to understand VSL, it is necessary to introduce a few terms used throughout the paper. These are in order:
\begin{itemize}
  \setlength\itemsep{0em}
  \item \textbf{\emph{World}:} {the workspace of the robot observable by the vision sensor. The \emph{world} includes objects being used during the learning task, and reconfigured by the human tutor and the robot.}
  \item \textbf{\emph{Frame}:} {a bounding box defining a rectangle in 2D space or a cuboid in 3D space. The size of the \emph{frame} can be fixed or variable. The maximum size of the \emph{frame} is equal to the size of the \emph{world}.}
  \item \textbf{\emph{Observation}:} {the captured context of the \emph{world} from a predefined viewpoint using a specific \emph{frame}. An \emph{observation} can be a 2D image or a 3D point cloud.}
  \item \textbf{\emph{Pre-action observation}:} {an \emph{observation} captured just before the action is executed. The robot searches for preconditions in the \emph{pre-action observations} before selecting and executing an action.}
  \item \textbf{\emph{Post-action observation}:} {an \emph{observation} captured just after the action is executed. The robot perceives the effects of the executed actions in the \emph{post-action observations}.}
\end{itemize}

The set of actions contains different primitive skills such as pick, place, push, or pull. VSL assumes that actions are known to the robot and the robot can execute each action when required. In Section~\ref{sec:symbolic}, VSL-SP extends VSL by teaching the primitive actions to the robot through demonstrations \citep{ahmadzadeh2015learning}.

\subsection{Problem Statement}
\label{subsec:problem}

Formally, we define visuospatial skill learning as a tuple
\begin{equation}
\mathcal{V} = \left \{ \mathcal{W} , \mathcal{O}, \mathcal{F}, \mathcal{A}, \mathcal{C}, \Pi, \phi \right \},
\end{equation}
\noindent where $\mathcal{W} \in \mathbb{R}^{m \times n}$ is a matrix which represents the context of the \emph{world} including the workspace and all objects ($\mathcal{W}_{D}$ and $\mathcal{W}_{R}$ refer to the \emph{world} during the demonstration and reproduction phases, respectively). $\mathcal{O}$ is a set of \emph{observation} dictionaries $\mathcal{O} = \left \{ \mathcal{O}^{Pre}, \mathcal{O}^{Post} \right \} $; $\mathcal{O}^{Pre}$ and $\mathcal{O}^{Post}$ are \emph{observation} dictionaries comprising a sequence of \emph{pre-action} and \emph{post-action observations}, respectively such that: $ \mathcal{O}^{Pre} = \langle \mathcal{O}^{Pre}(1),\mathcal{O}^{Pre}(2), \ldots , \mathcal{O}^{Pre}(\eta) \rangle $, and $\mathcal{O}^{Post} = \langle \mathcal{O}^{Post}(1),\mathcal{O}^{Post}(2), \ldots , \mathcal{O}^{Post}(\eta) \rangle $, where $\eta$ is the number of operations performed by the tutor during the demonstration \index{demonstration} phase. As a consequence, $\mathcal{O}^{Pre}(i)$ represents the \emph{pre-action observation} captured during the $i^{\mathrm{th}}$ operation. %
$\mathcal{F} \in \mathbb{R}^{m \times n}$ is an \emph{observation frame} used to capture \emph{observations}. $\mathcal{A}$ is a set of primitive actions defined in the learning \index{learning} task (e.g. pick). $\mathcal{C}$ is a set of \emph{constraint} dictionaries $\mathcal{C} = \left \{ \mathcal{C}^{Pre}, \mathcal{C}^{Post} \right \} $; $ \mathcal{C}^{Pre}$ and $ \mathcal{C}^{Post}$ are \emph{constraint} dictionaries comprising a sequence of \emph{pre-action}, and \emph{post-action constraints}, respectively. $\Pi$ is a policy or an ordered action sequence extracted from demonstrations.  \index{demonstration} $\phi$ is a vector containing extracted features from \emph{observations} (e.g. SIFT features).

\IncMargin{0.5em}
\begin{algorithm*}[ht]
\SetAlFnt{\small}
\DontPrintSemicolon
\SetKwInOut{Input}{Input}\SetKwInOut{Output}{Output}
\Input {$\left \{ \mathcal{W} , \mathcal{F} , \mathcal{A} \right \} $}
\Output{$\left \{ \mathcal{O}, \mathcal{P} , \Pi ,\mathcal{C}, \mathcal{B}, \phi \right \} $}
\textbf{Initialization:} \small{place objects in the world randomly}\;\label{alg:line1}
    $\mathcal{L}$=\textsf{detectLandmarks}$(\mathcal{W})$\; \label{alg:line2}
    $i = 1$\;
    \tcp{\small{Part I : Demonstration}}
    \For {each Action}
    {\label{alg:line4}
         $\mathcal{O}^{Pre}_{i} $ = \textsf{getPreActionObs}$(\mathcal{W}_{D},\mathcal{F}_{D})$\; \label{alg:line5}
         $\mathcal{O}^{Post}_{i} $ = \textsf{getPostActionObs}$(\mathcal{W}_{D},\mathcal{F}_{D})$\; \label{alg:line6}
          $[\mathcal{B}_{i},\mathcal{P}^{Pre}_{i},\mathcal{P}^{Post}_{i},\phi_{i}]$ = \textsf{getObject}$(\mathcal{O}^{Pre}_{i},\mathcal{O}^{Post}_{i})$\;\label{alg:line7}
         $[\mathcal{C}^{Pre}_{i},\mathcal{C}^{Post}_{i}]$ = \textsf{getConstraint}$(\mathcal{B}_{i},\mathcal{P}^{Pre}_{i},\mathcal{P}^{Post}_{i},\mathcal{L})$\; \label{alg:line8}
         $\Pi_{i} $ = \textsf{getAction}$(\mathcal{A},\mathcal{C}^{Pre}_{i},\mathcal{C}^{Post}_{i})$\; \label{alg:line9}
         $i = i + 1$\;
    }

      \tcp{\small{Part II : Reproduction}}
    \For{$j = 1 \; \textbf{to} \; i$}
    {
        $\mathcal{P}^{*Pre}_{j}$ = \textsf{findBestMatch} $(\mathcal{W}_{R},\mathcal{O}^{Pre}_{j},\phi_{j},\mathcal{C}^{Pre}_{j},\mathcal{L})$\; \label{alg:line13}
        $\mathcal{P}^{*Post}_{j}$ = \textsf{findBestMatch} $(\mathcal{W}_{R},\mathcal{O}^{Post}_{j},\phi_{j},\mathcal{C}^{Post}_{j},\mathcal{L})$\; \label{alg:line14}
        \textsf{executeAction}$(\mathcal{P}^{*Pre}_{j},\mathcal{P}^{*Post}_{j},\Pi_{j})$\; \label{alg:line15}

     }
\caption{Pseudo-code for VSL. \label{alg1}}
\end{algorithm*}
\DecMargin{0.5em}

\subsection{Methodology}
\label{subsec:method}

Pseudo-code of VSL is sketched in Algorithm~\ref{alg1}. At the beginning  (line~\ref{alg:line1}), objects are randomly placed in the \emph{world} ($\mathcal{W}_{D}$) and the size of the \emph{frame} ($\mathcal{F}_{D}$) is equal to the size of the \emph{world} ($\mathcal{W}_{D}$). We assume that the robot has knowledge of a set of primitive actions $\mathcal{A}$ and it is capable of executing them. For instance, the robot is capable of moving towards a desired given pose and execute a pick action. In Section~\ref{sec:symbolic}, we explain how a primitive action library can be built.

In some tasks, to specify different spatial concepts, the tutor can use a landmark. For instance, a vertical borderline can be used to divide the workspace into two areas illustrating right and left zones. A landmark can be either static or dynamic. A static landmark is fixed with respect to the \emph{world} during both demonstration and reproduction. A dynamic landmark, however, can be replaced by the tutor before the reproduction phase starts. Both types of landmarks are shown in this paper. In case that any landmarks (e.g. label, borderline) are used in the demonstration phase, the robot should be able to detect them in the \emph{world} (line~\ref{alg:line2}). However, we assume that the robot cannot manipulate a landmark.

During the demonstration phase, VSL captures one \emph{pre-action observation} ($\mathcal{O}^{Pre}$) and one \emph{post-action observation} ($\mathcal{O}^{Post}$) for each operation executed by the tutor using the specified \emph{frame} ($\mathcal{F}_{D}$) (lines~\ref{alg:line5},and \ref{alg:line6}). We assume that during each operation only one object can be affected by an action. The \emph{pre-action} and \emph{post-action observations} are used to detect the object on which the action is executed, the pose of the object before and after the action execution ($\mathcal{P}^{Pre},\mathcal{P}^{Post}$) (line~\ref{alg:line7}). VSL is capable of detecting objects during the demonstration process without having any \textit{a priori} knowledge in advance. Practically, this can be achieved using simple image processing techniques such as image-subtraction and thresholding applied on a couple of raw \emph{observations}.
For each detected object, a symbolic representation ($\mathcal{B}$) is created (line~\ref{alg:line7}). Such symbolic representation of an object can be used in case the reproduction phase of VSL is integrated with a high-level symbolic planner as shown in Section~\ref{sec:symbolic}. In addition, VSL extracts a feature vector ($\phi$) for each detected object (line~\ref{alg:line7}). In order to extract $\phi$, any feature extracting method can be used. In this Section, we use Scale Invariant Feature Transform (SIFT) algorithm \citep{lowe2004distinctive}. SIFT is one of the most popular feature-based methods which is able to detect and describe local features that are invariant to scaling and rotation.

Pose vectors and the detected landmarks ($\mathcal{L}$) are used to identify preconditions and effects of the executed action through spatial reasoning (line~\ref{alg:line8}). For instance, if $\mathcal{P}^{Pre}$ is above a horizontal borderline and $\mathcal{P}^{Post}$ is below the line, the precondition of the action is that the object is above the line and the effect of the execution of the action is that the object is below the line. By observing the predicates of an executed action, the action itself can be identified from the set of actions $\mathcal{A}$ (line~\ref{alg:line9}). The sequence of identified actions is then stored in a policy vector $\Pi$.

During the reproduction phase, VSL is able to execute the learned sequence of actions independent of any external action planner  \citep{ahmadzadeh2013interactive,ahmadzadeh2013visuospatial}. In such case, VSL observes the new \emph{world} ($\mathcal{W}_{R}$) in which the objects are replaced randomly. Comparing the recorded \emph{pre-} and \emph{post-action observations}, with the new \emph{world}, VSL detects the best matches for each objects, considering the extracted features, new location of the landmarks, and the extracted constraints from the learning phase (lines ~\ref{alg:line13} and ~\ref{alg:line14}). Although the \textsf{findBestMatch} function can use any metric (e.g. window search method) to find the best matching \emph{observation}, to be consistent, in all of our experiments in this Section, we use SIFT. Afterward, we apply Random Sample Consensus method (RANSAC) in order to estimate the transformation matrix $\mathbf{T}_{\mathrm{sift}}$ from the set of matches. A normalization constant is used to have a unique normalized representation and avoiding unnatural interpolation results. The normalized matrix $\alpha\mathbf{T}_{\mathrm{sift}}$ can be decomposed into simple transformation elements,
\begin{equation}
\alpha\mathbf{T}_{\mathrm{sift}} = \mathbf{T}\mathbf{R}_{\theta}\mathbf{R}_{-\phi}\mathbf{S}_{v}\mathbf{R}_{\phi}\mathbf{P}, \nonumber
\end{equation}

\noindent where $\mathbf{R}_{\pm\phi}$ are rotation matrices to align the axis for horizontal and vertical scaling of $\mathbf{S}_{v}$; $\mathbf{R}_{\theta}$ is another rotation matrix to orientate the shape into its final orientation; $\mathbf{T}$ is a translation matrix; and lastly $\mathbf{P}$ is a pure projective matrix. An affine matrix, $\mathbf{T}_{A}$, that is the remainder of $\alpha\mathbf{T}$ by extracting $\mathbf{P}$ can be calculated as $\mathbf{T}_{A}=\alpha\mathbf{T}\mathbf{P}^{-1}$. $\mathbf{T}$ is extracted by taking the $3^{\mathrm{rd}}$ column of $\mathbf{T}_{A}$ and $\mathbf{A} \in \mathbb{R}^{2\times2}$, is the remainder of $\mathbf{T}_{A}$. $\mathbf{A}$ can be further decomposed using SVD such that, $\mathbf{A} = \mathbf{UDV}^{T}$, where $\mathbf{D}$ is a diagonal matrix, and $\mathbf{U}$ and $\mathbf{V}$ are orthogonal matrices. Finally, we can calculate
\begin{align}
\mathbf{S}_{v} = \left[
\begin{matrix}
  \mathbf{D} & \textbf{0} \\
  \textbf{0} & 1
\end{matrix}
\right],
\mathbf{R}_{\theta} = \left[
\begin{matrix}
  \mathbf{UV}^{T} & \textbf{0} \\
  \textbf{0} & 1
\end{matrix}
\right],
\mathbf{R}_{\phi} = \left[
\begin{matrix}
  \mathbf{V}^{T} & \textbf{0} \\
  \textbf{0} & 1
\end{matrix}
\right]. \nonumber
\end{align}

In a pick-and-place task, since $\mathbf{R}_{\theta} $ is calculated for both pick and place operations ($\mathbf{R}_{\theta}^{Pick}, \mathbf{R}_{\theta}^{Place}$), the pick and place rotation angles of the objects are extracted as follows:
\begin{eqnarray}
\theta^{Pick}  &= \arctan{\left(\frac{\mathbf{R}_{\theta}^{Pick}(2,2)}{\mathbf{R}_{\theta}^{Pick}(2,1)} \right)} \label{eq:angle1} \\
\theta^{Place}  &= \arctan{ \left(\frac{\mathbf{R}_{\theta}^{Place}(2,2)}{\mathbf{R}_{\theta}^{Place}(2,1)}\right)}. \label{eq:angle2}
\end{eqnarray}

\begin{figure}[ht] 
      \centering
      \includegraphics[width=\columnwidth]{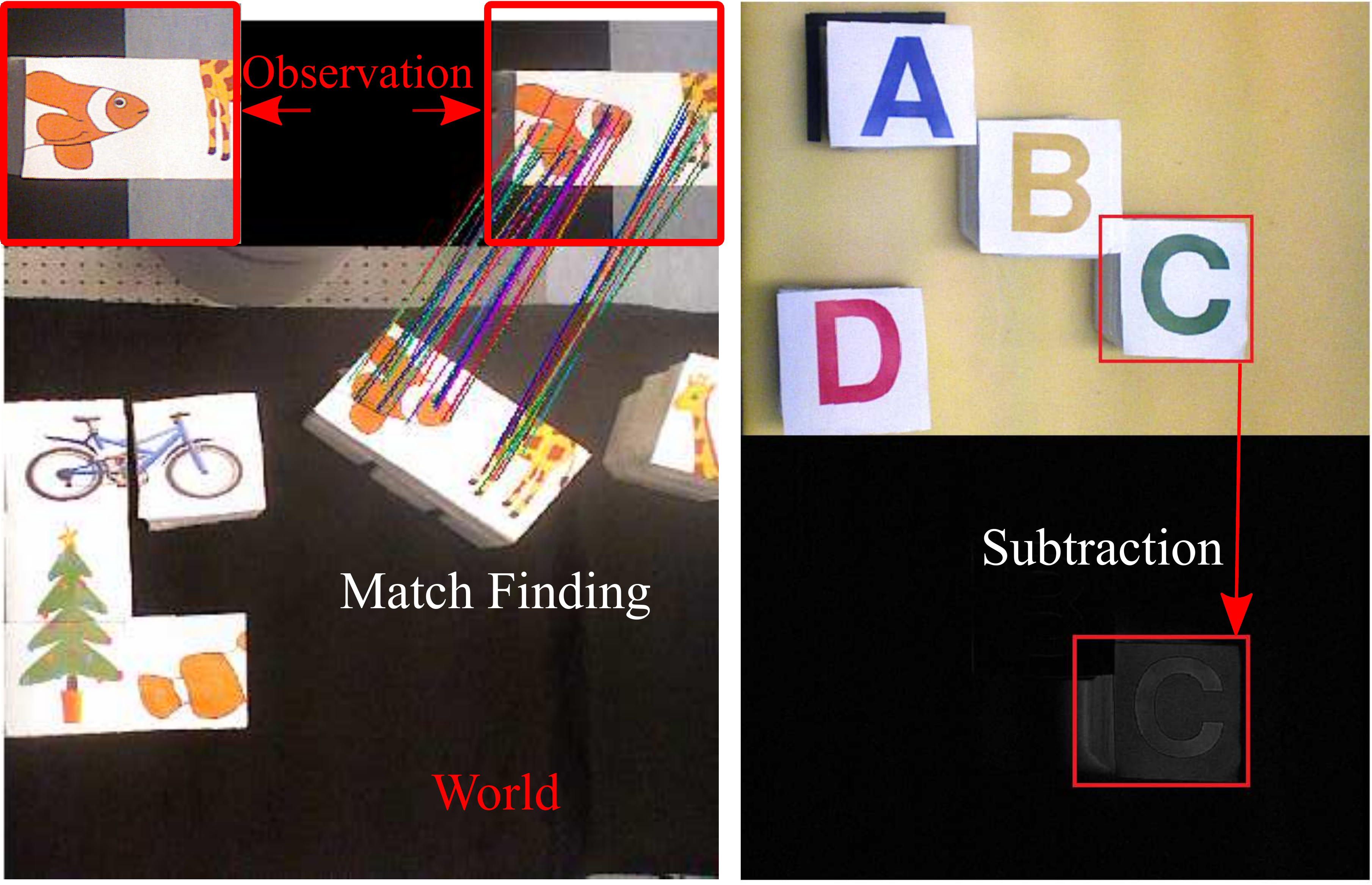} 
      \caption{The result of the image subtracting and thresholding for a place action (right), match finding result using SIFT in the $4^{\mathrm{th}}$ operation during reproduction of the Domino task (left).} 
      \label{fig:sift}
\end{figure}

VSL relies on vision, which might be obstructed by other objects, by the tutor's body, or during the reproduction by the robot's arm. Therefore, for the physical implementation of the VSL approach special care needs to be taken to avoid such obstructions. Figure~\ref{fig:sift} shows the result of match finding using SIFT applied to an \emph{observation} and a new \emph{world}.

After finding the best match, the Algorithm extracts the pose of the object before and after action execution, $\mathcal{P}_{j}^{*Pre}$, $\mathcal{P}_{j}^{*Post}$ (lines~\ref{alg:line13}~and~\ref{alg:line14}). Finally, an action is selected from the policy $\Pi$ and together with pre and post poses is sent to the \textsf{executeAction} function (line \ref{alg:line15}). This function selects the $\mathcal{A}_{j}$ primitive action.

\begin{figure}[ht] 
 \centering
  \subfloat[A full cycle of spatial trajectory generated for a pick-and-place operation.]%
    {
    \includegraphics[trim=0cm 0cm 0cm 0cm, width=0.9\columnwidth]{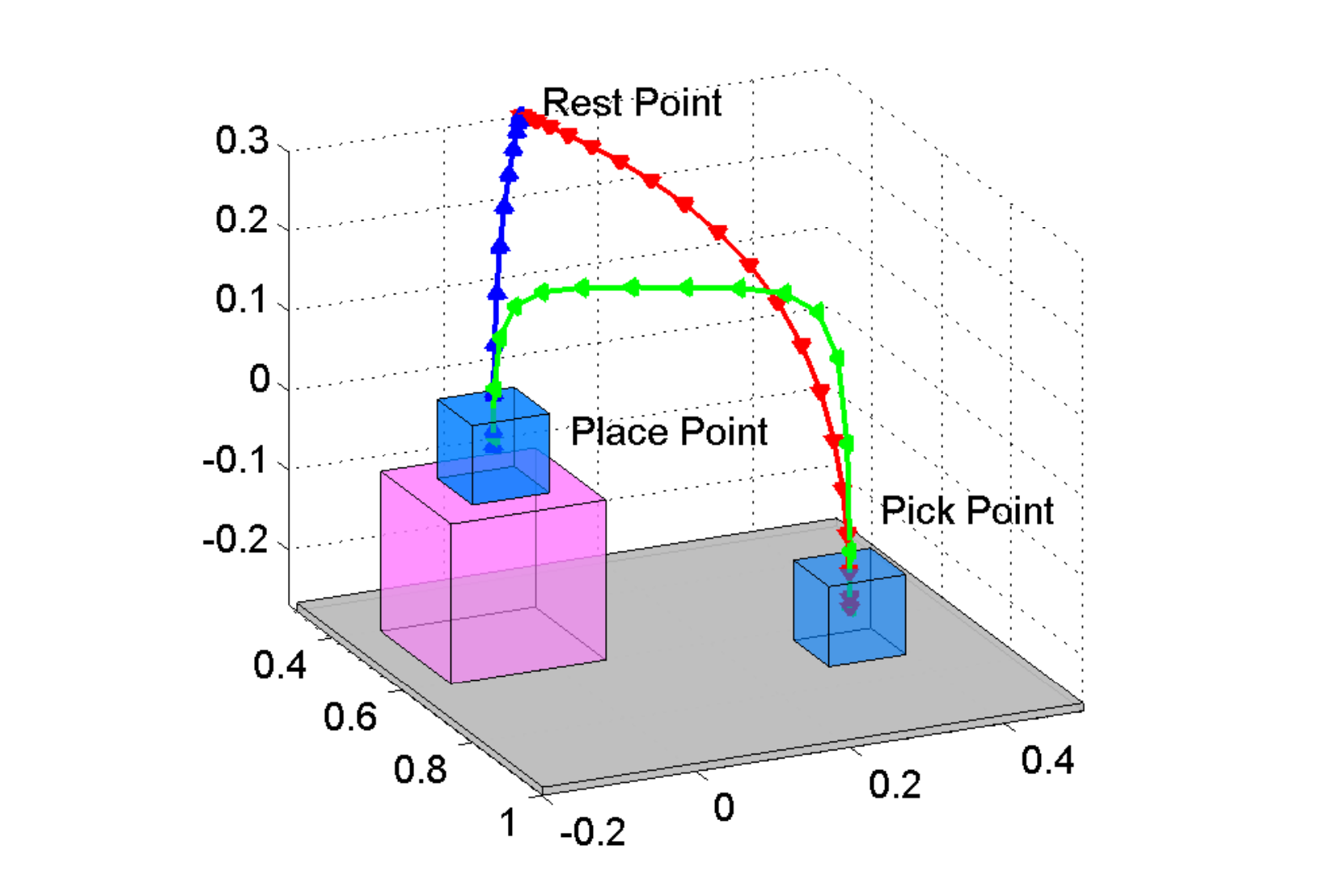} 
    \label{fig:trajXYZ2}
    }
    \\
  \subfloat[The generated angle of rotation, $\theta$, for the robot's hand.]
    {
    \includegraphics[trim=0cm 0 0 0,width=0.8\columnwidth]{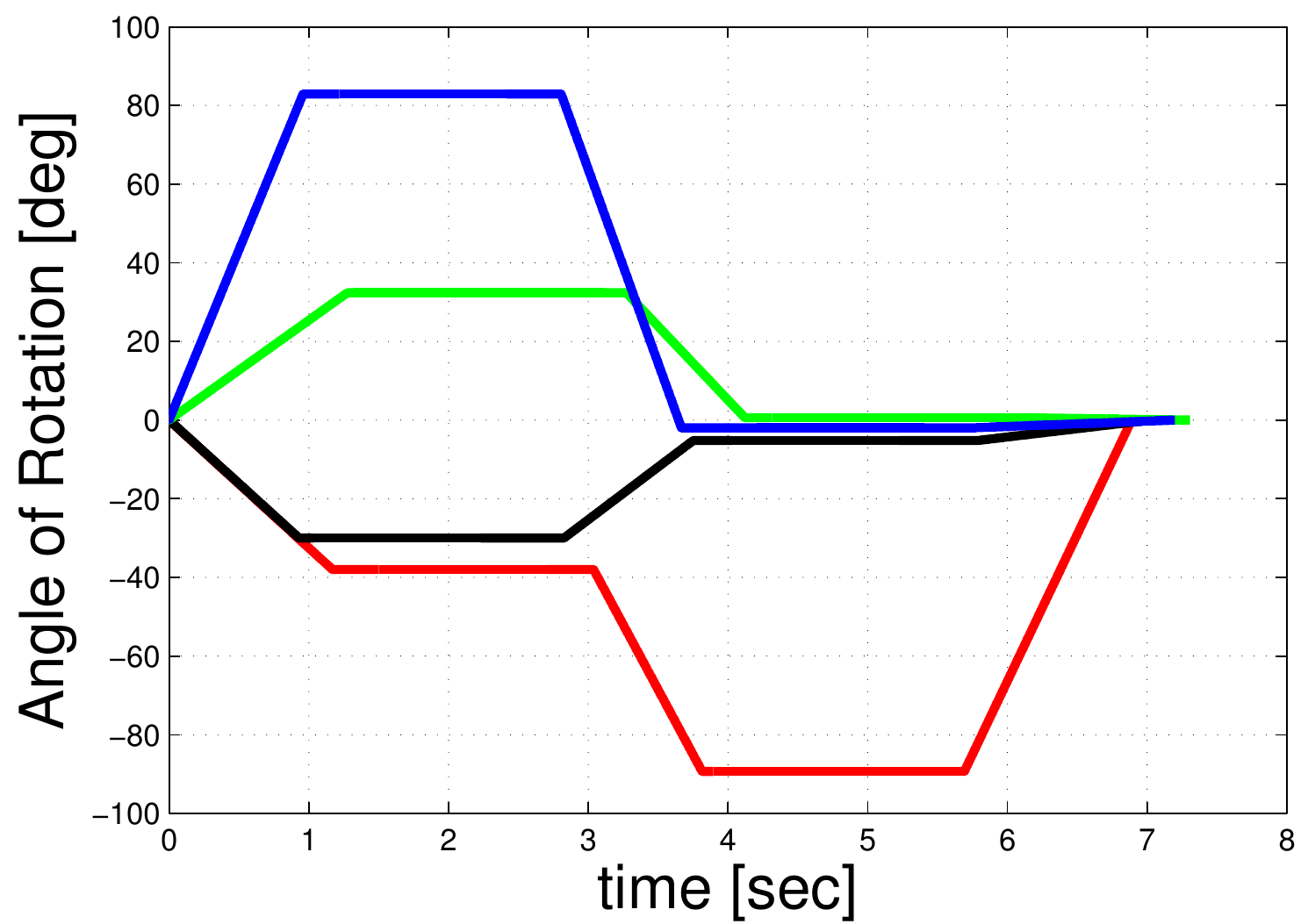}
    \label{fig:trajROT}
    }
 \caption{The generated trajectory cycle including position and orientation profiles for a spatial pick-and-place task.}
\label{fig:traj1}
\end{figure}

To perform a pick-and-place operation, the robot must execute a set of primitive actions consisting of reaching, grasping, relocating, and releasing. Either one of the reaching and relocating actions are trajectory-based skills which can either be manually programmed to the system or a tutor can teach them to the robot for instance through learning by demonstration technique~\citep{ahmadzadeh2015learning} as shown in Section~\ref{sec:symbolic}. In the original VSL~\citep{ahmadzadeh2013visuospatial}, a simple trajectory generation strategy was employed that we explain here. The extracted pick and place points ($\mathcal{P}_{j}^{*Pick}, \mathcal{P}_{j}^{*Place}$) together with the corresponding rotation angles ($\theta_{j}^{Pick}, \theta_{j}^{Place}$), are used to generate a trajectory for the corresponding operation. For each pick-and-place operation, the desired Cartesian trajectory of the end-effector is a cyclic movement between three key points: a rest point, a pick point, and a place point. Figure~\ref{fig:trajXYZ2} illustrates a complete trajectory cycle generated for a pick-and-place operation. Also, four different profiles of rotation angles are depicted in Figure~\ref{fig:trajROT}. The robot starts from the rest point while the rotation angle is equal to zero and moves smoothly along the red curve towards the pick point, $\mathcal{P}_{j}^{*Pick}$. During this movement, the robot's hand rotates to satisfy the pick rotation angle, $\theta_{j}^{Pick}$, according to the rotation angle profile. Then the robot picks up an object, relocates it along the green curve to the place point, $\mathcal{P}_{j}^{*Place}$, while the hand is rotating to meet the place rotation angle, $\theta_{j}^{Place}$. Then, the robot places the object in the place point and finally moves back along the blue curve to the rest point. In order to form each trajectory, initial conditions (i.e. initial positions and velocities) and a specific duration must be defined. Thereby, a geometric path is defined which can be expressed in the parametric form of the following equations:%
\begin{align}
{p}_{x} &=& {a}_{3}{s}^{3}+{a}_{2}{s}^{2}+{a}_{1}{s}+{a}_{0} \label{eq:traj1} \\
{p}_{y} &=& {b}_{3}{s}^{3}+{b}_{2}{s}^{2}+{b}_{1}{s}+{b}_{0} \label{eq:traj2} \\
{p}_{z} &=& h[1-|{({\tanh}^{-1}({h}_{0}(s-0.5)))}^{\kappa}|] , \label{eq:traj3}
\end{align}
\noindent where, $s$ is a function of time $t$, ($s = s(t)$), ${p}_{x}={p}_{x}(s)$, ${p}_{y}={p}_{y}(s)$, and ${p}_{z}={p}_{z}(s)$ are the 3D elements of the geometric spatial path; The $a_{i}$ and $b_{i}$ coefficients  in \eqref{eq:traj1} and \eqref{eq:traj2} are calculated using the initial and final conditions. $\kappa$ in \eqref{eq:traj3} is the curvature, $h_{0}$ and $h$ are the initial height and the height of the curve in the middle point of the path respectively. $h$ and $h_{0}$ can be either provided by the tutor or detected through depth information provided by an RGB-D sensor. In addition, the time is smoothly distributed with a $3^{\mathrm{rd}}$ order polynomial between the $t_{start}$ and $t_{final}$ which both are instructed by the user. Moreover, to generate a rotation angle trajectory for the robot's hand, a trapezoidal profile is used together with the $\theta^{Pick}$ and $\theta^{Place}$ calculated in \eqref{eq:angle1} and \eqref{eq:angle2}. As shown in Figure~\ref{fig:trajXYZ2}, the trajectory generation module can also deal with objects placed in different heights (different $z$-axis levels).

It is worth noting that to reproduce the task with more general capabilities, a generic symbolic planner can be utilized instead of the reproduction part of the Algorithm. Such extension is described in Section~\ref{sec:symbolic}.

Another vital element for implementing a pick-and-place operation is the grasping strategy. There are many elaborate ways to do grasp synthesis for known or unknown objects~\citep{bohg2014data,su2012robust}. Since the problem of grasping is not the main focus of our research, we implement a simple but efficient grasping method using the torque sensor of the robots used in the experiments. The grasping position is calculated using the center of the corresponding \emph{pre-action} observation. The grasping module firstly opens all the fingers and closes them after the hand is located above the desired object. The fingers stop closing when the measured torque is more than a pre-defined threshold value. In addition, by estimating a bounding box for the target observation, the values are used to decide which axis is more convenient for grasping.

\subsection{Simulation}
\label{subsec:simulation}
\begin{figure}[ht] 
    \centering
    \subfloat[Demonstration]
    {
        \includegraphics[width=0.8\columnwidth]{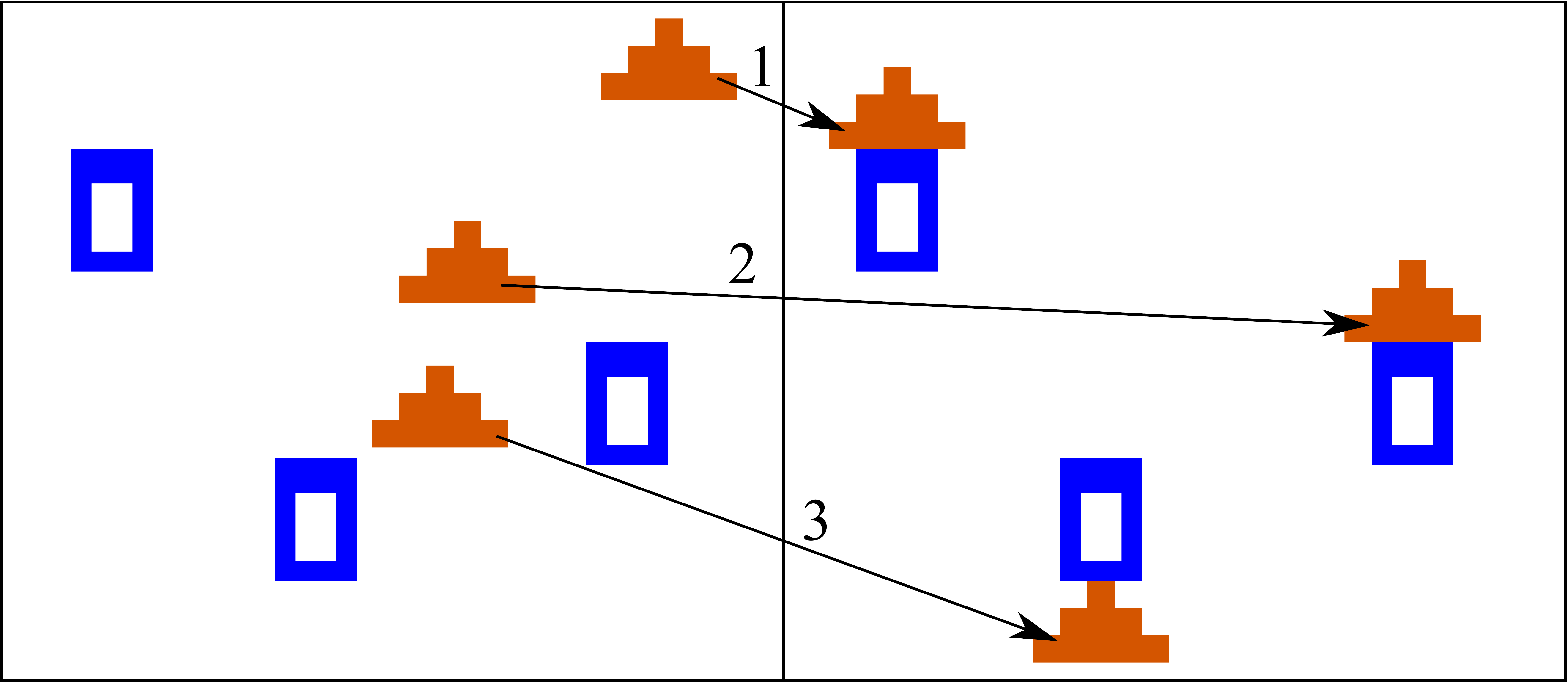} 
        \label{fig:3HD}
    }
    \\
   \subfloat[Three steps of adaptive frame-size ]
    {
        \includegraphics[width=0.8\columnwidth]{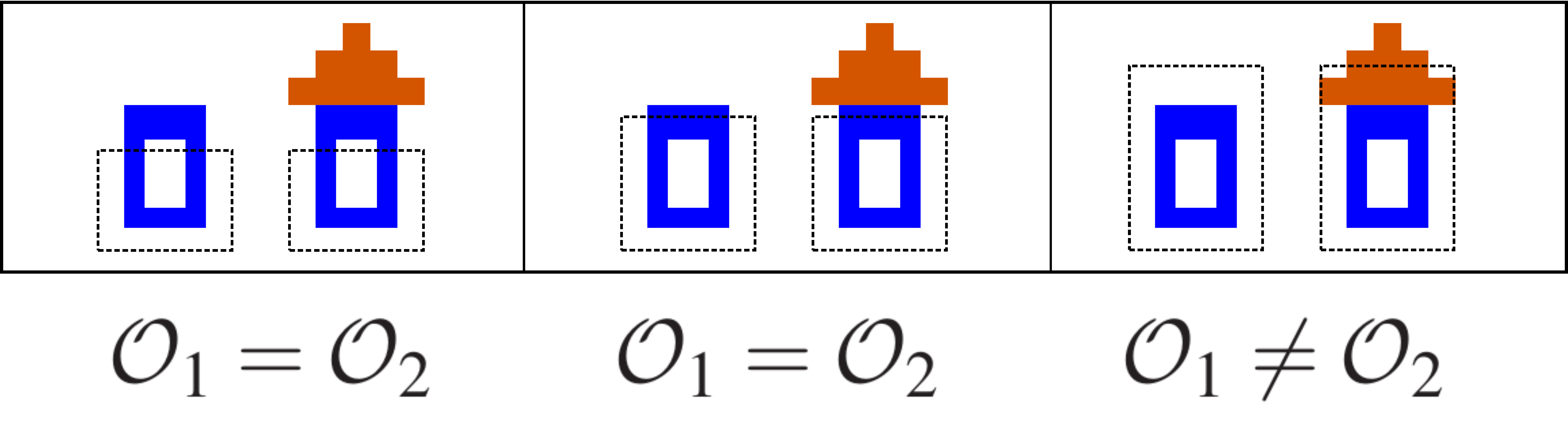} 
        \label{fig:3AF}
    }
    \\
    \subfloat[Reproduction]
    {
        \includegraphics[width=0.8\columnwidth]{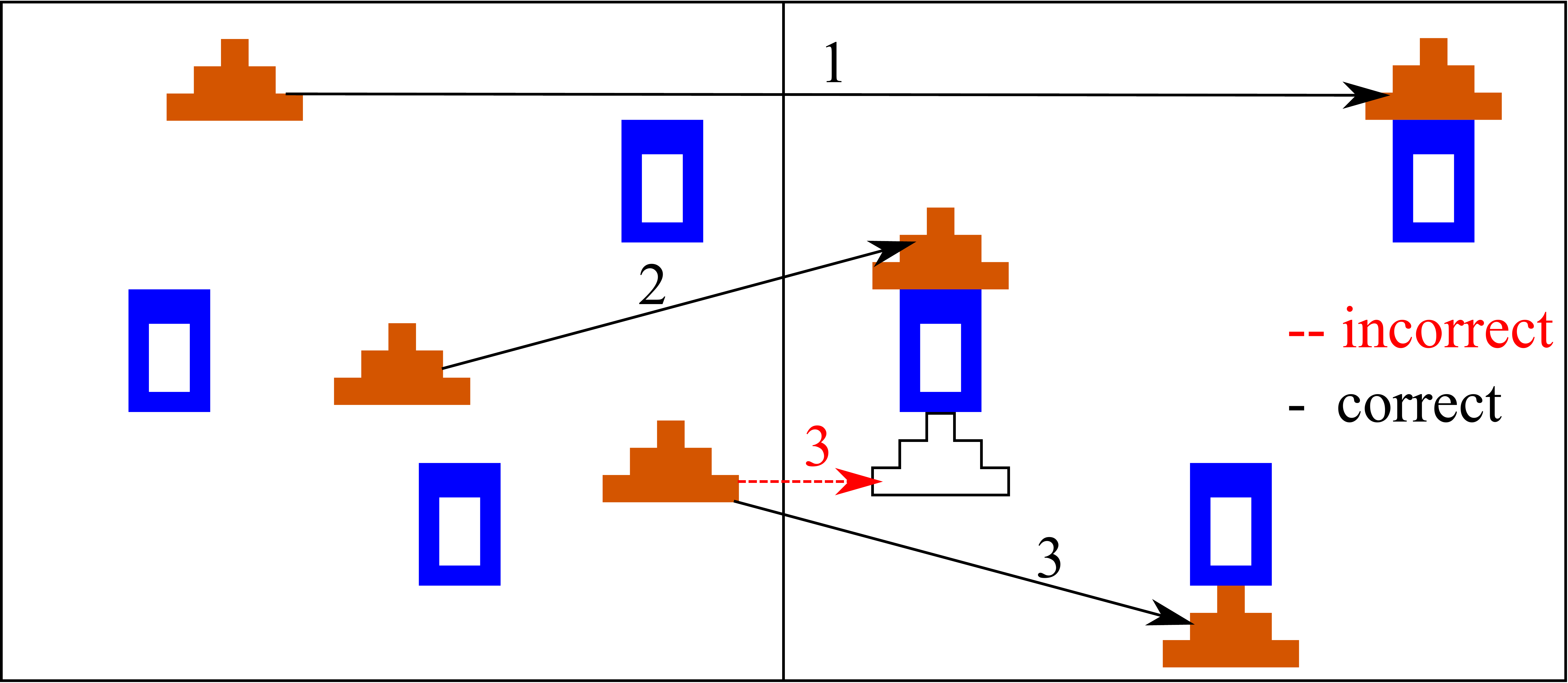} 
        \label{fig:3HR}
    }
    \caption{Roof placement simulated experiment to illustrate VSL's capability of disambiguation of multiple alternative matches.}
    \label{fig:3Houses}
\end{figure}

In this Section, a simulated experiment is designed to gain an understanding of how VSL operates and to show the main idea of VSL without dealing with practical limitations and implementation difficulties. For this simulation, as shown in Figure~\ref{fig:3Houses}, a set of 2D objects is made which the tutor can manipulate and assemble them on an empty workspace using keyboard or mouse. Each operation consists of a \emph{pick} and a \emph{place} action, which are executed by holding and releasing a mouse button.

In this task, the \emph{world} includes two sets, each containing three visually identical objects (i.e. three blue `house bodies' and three brown `roofs'). As it can be seen in Figure~\ref{fig:3HD}, the tutor selects the `roof' objects arbitrarily and places them on top of the `bodies'. However, in the $3^{\mathrm{rd}}$ operation, the tutor intentionally puts the `roof' at the bottom of the `house body'. The goal of this experiment is to show the VSL's capability of disambiguation of multiple alternative matches. If the algorithm uses a fixed search \emph{frame} ($\mathcal{F}_{R}$) that is smaller than the size of the `bodies' (i.e. blue objects in the \emph{world}), then, as shown in the first and second sub-figures~\ref{fig:3AF}, the two captured observations can become equivalent (i.e. $\mathcal{O}_1 = \mathcal{O}_2$) and the $3^{\mathrm{rd}}$ operation might be performed incorrectly (see the incorrect reproduction in Figure~\ref{fig:3HR}). The reason is that, due to the size of the \emph{frame}, the system perceives a section of the \emph{world} not bigger than the size of a `house body'. The system is not aware that an object is already assembled on the top and it will select the first matching \emph{pre-place} position to place the last object there. To resolve this problem we use adaptive size \emph{frames} during the match finding process in which the size of the \emph{frame} starts from the smallest feasible size and grows up to the size of the \emph{world}. The function \textsf{findBestMatch} in Algorithm~\ref{alg1}, is responsible for creating and changing the size of the \emph{frame} adaptively in each step. This technique helps the robot to resolve the ambiguity issue by adding more context inside the observation, which in effect narrows down the possible matches and leaves only a single matching observation. Figure~\ref{fig:3HR} shows the sequence of reproduction including correct and incorrect operations.

\subsection{Real-world Experiments}
\label{subsec:realexp}

\begin{figure}[ht]
      \centering
      \includegraphics[width=0.9\columnwidth]{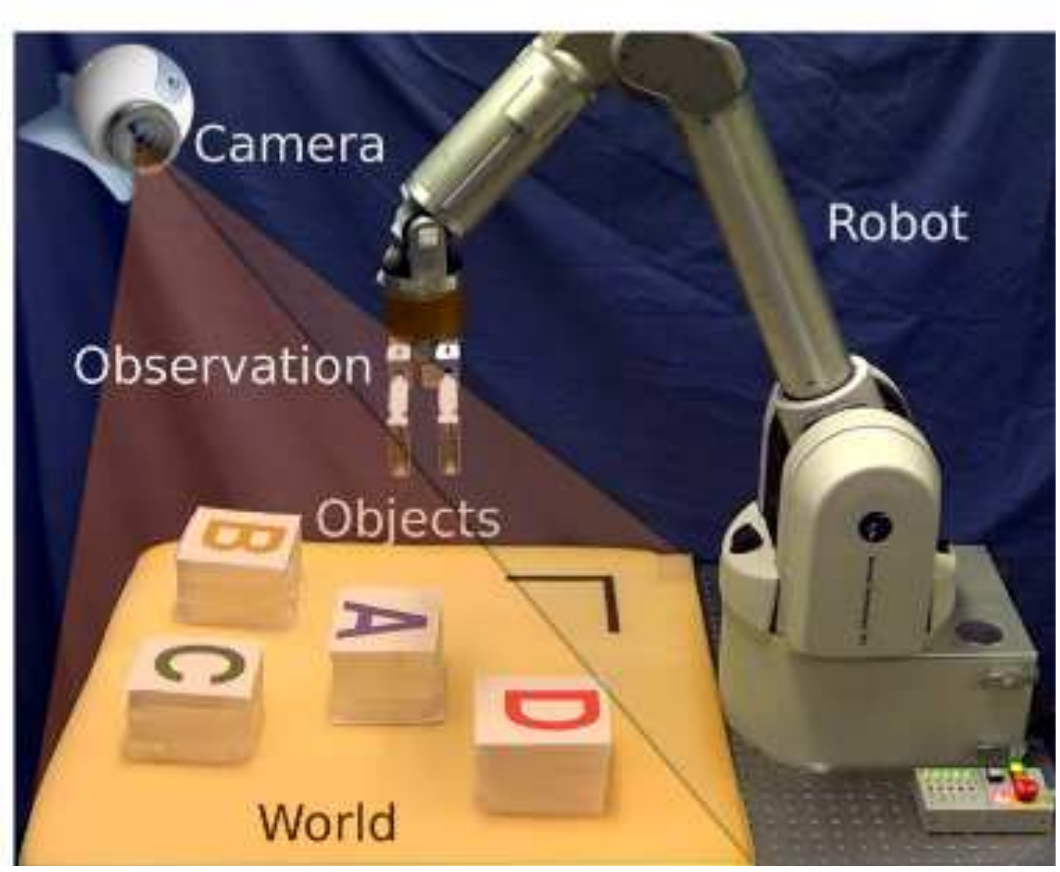} 
      \caption{The experimental setup for a VSL task.}
      \label{fig:setup}
\end{figure}

In this Section, five real-world experiments are described. As it can be seen in Figure~\ref{fig:setup}, the setup for all the experiments consists of a torque-controlled $7$~DOF Barrett WAM robotic arm equipped with a $3$-finger Barrett Hand, a tabletop working area, a set of objects, and a CCD camera mounted above the workspace (not necessarily perpendicular to it). The resolution of the captured images is $1280 \times 960$ pixels. In all the conducted experiments, the robot learns simple object manipulation tasks including pick-and-place actions. In order to perform a pick-and-place operation, the extracted pick and place poses are used to make a cyclic trajectory as explained in Section~\ref{subsec:method}. In the demonstration phase, the size of the \emph{frame} for the \emph{pre-pick observation} is set equal to the size of the biggest object in the \emph{world}, and the size of the \emph{frame} for the \emph{pre-place observation} two or three times bigger than the size of the biggest objects in the \emph{world}. In the reproduction phase, the size of the \emph{frame} is set equal to the size of the \emph{world}. Note that, in all the illustrations, the straight and curved arrows are used just to show the sequence of operations, not the actual trajectories for performing the movements. Table~\ref{Table1} summarizes different capabilities of VSL emphasized in each task.

\begin{table*}[ht]
\caption{Capabilities of VSL illustrated in each real-world experiment.}
\label{Table1}
\small
\centering
\begin{tabular}{|lr|c|c|c|c|c|}
\hline
    {\multirow{2}{*}{\textbf{Capability}}} &   \multirow{2}{*}{\rotatebox{90}{\textbf{Task}}}     & {Animal}  & {Alphabet}  & {Tower of} & {Animals vs. }  & {\multirow{2}{*}{Domino}}\\
    & & {{Puzzle}} & {{Ordering}} & {{Hanoi}} & {{Machines}} & \\ \hline
    \multicolumn{2}{|l|}{Relative positioning} & $\checkmark$ & $\checkmark$ & $\checkmark$ & - & $\checkmark$ \\ \hline
    \multicolumn{2}{|l|}{Absolute positioning} & - & $\checkmark$ & $\checkmark$ & - & - \\ \hline
    \multicolumn{2}{|l|}{Classification} & - & - & - & $\checkmark$ & - \\ \hline
    \multicolumn{2}{|l|}{Turn-taking} & - & - & $\checkmark$ & $\checkmark$ & $\checkmark$ \\ \hline
    \multicolumn{2}{|l|}{User intervention} & - & - & $\checkmark$ & $\checkmark$ & $\checkmark$ \\ \hline
   \end{tabular}
\end{table*}%

In order to test the repeatability of VSL and to identify the possible factors of failure, the captured observations from the real-world experiments were used while excluding the robot from the loop. All other parts of the loop were kept intact and each experiment was repeated three times. The result shows that less than 5\% of pick-and-place operations failed. The main failure factor is the match finding error which can be resolved by adjusting the parameters of SIFT-RANSAC or using alternative match finding algorithms. The noise in the images and the occlusion of the objects can be listed as two other potential factors of failure. Despite the fact that VSL is scale-invariant, color-invariant, and view-invariant, it has some limitations. For instance, if the tutor accidentally moves one object while operating another, the algorithm may fail to find a pick/place position. One possible solution is to combine classification techniques together with the image subtraction and thresholding techniques to detect multi-object movements. Such extension is out of the scope of the presented work.

\subsubsection*{Alphabet Ordering}
\label{subsubsec:task1alphabet}

\begin{figure}[ht]
      \centering
      \subfloat[The sequence of the operations in the demonstration phase by the tutor]
      {
        \includegraphics[width=\columnwidth]{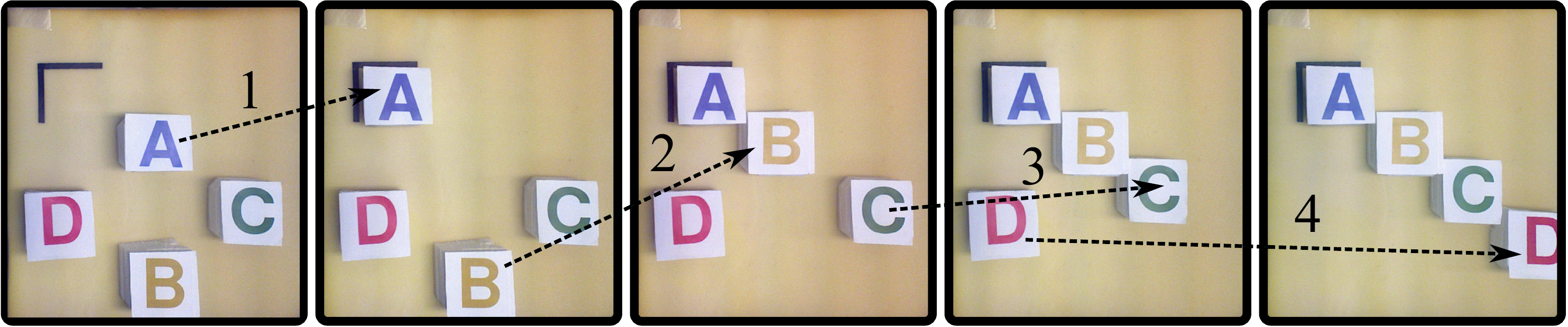} 
        \label{fig:ABCD_demo}
      }
      \qquad
      \subfloat[The sequence of the operations in the reproduction phase by the robot]
      {
        \includegraphics[width=\columnwidth]{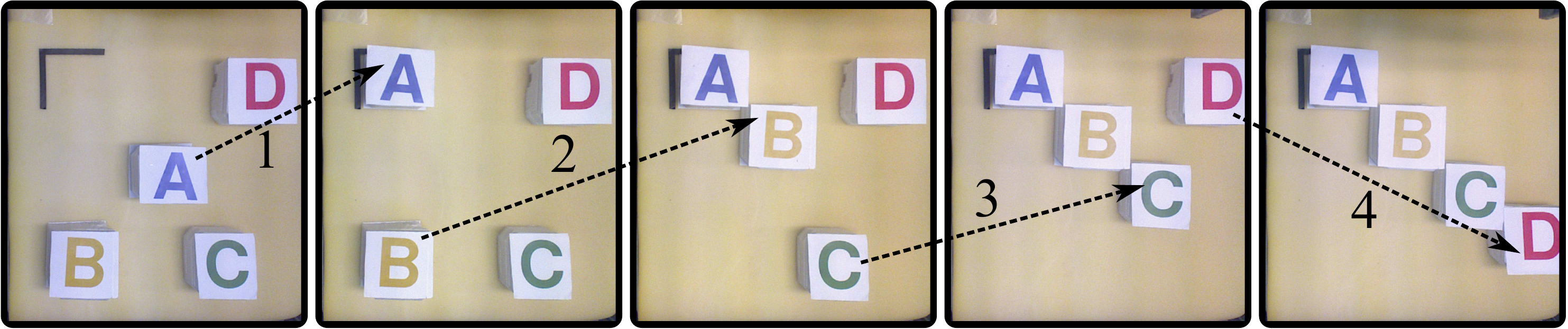} 
        \label{fig:ABCD_repro}
      }
      \caption{Alphabet ordering. The initial configuration of the objects in the \emph{world} is different in \ref{fig:ABCD_demo} and \ref{fig:ABCD_repro}. Black arrows show the operations.}
\end{figure}

In the first task, the \emph{world} includes four cubic objects labeled with $A$, $B$, $C$, and $D$ letters. The \emph{world} also includes a static right angle baseline as a landmark ($\mathcal{L}$). The goal is to reconfigure and sort the set of objects with respect to the baseline according to a previous demonstration. As reported in Table~\ref{Table1}, this task emphasizes VSL's capability of the relative positioning of an object with respect to other surrounding objects (a visuospatial skill). This is achieved through the use of visual \emph{observations} capturing both the object of interest and its surrounding objects (i.e. its context).
In addition, the baseline is provided to show the capability of absolute positioning of VSL. It shows the fact that we can teach the robot to attain absolute positioning of objects without defining any explicit \emph{a priori} knowledge. Figure~\ref{fig:ABCD_demo} shows the sequence of operations in the demonstration phase. Recording \emph{pre-pick} and \emph{pre-place observations}, the robot learns the sequence of operations. Figure~\ref{fig:ABCD_repro} shows the sequence of operations produced by VSL starting from a novel \emph{world} (i.e. a new initial configuration) achieved by randomizing objects locations in the \emph{world}. The accompanying video shows the execution of this task~\citep{iros2013video}.

\subsubsection*{Animal Puzzle}
\label{subsubsec:task2animal}

\begin{figure}[ht]
      \centering
      \subfloat[The sequence of the operations in the demonstration phase by the tutor]
      {
      \includegraphics[width=\columnwidth]{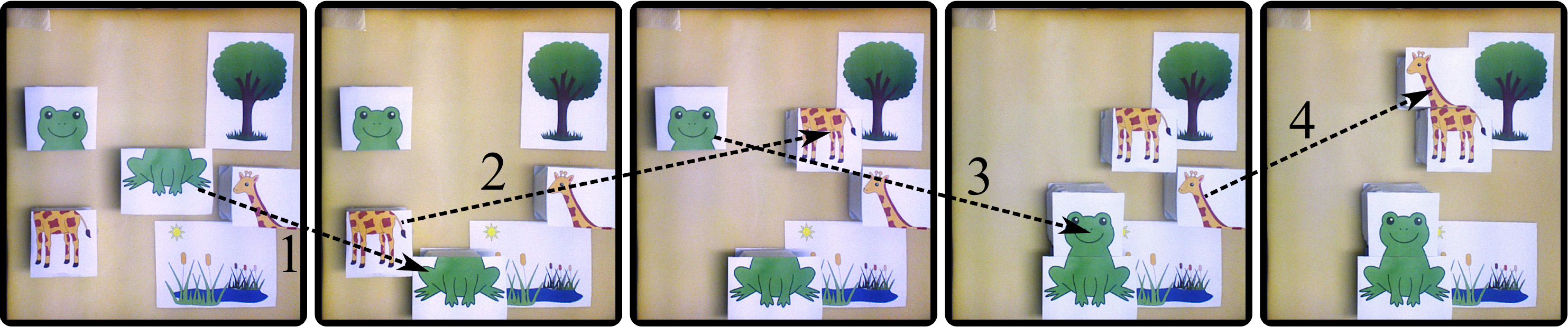} 
        \label{fig:Animal_demo}
      }
      \\
      \subfloat[The sequence of the operations in the reproduction phase by the robot]
      {
        \includegraphics[width=\columnwidth]{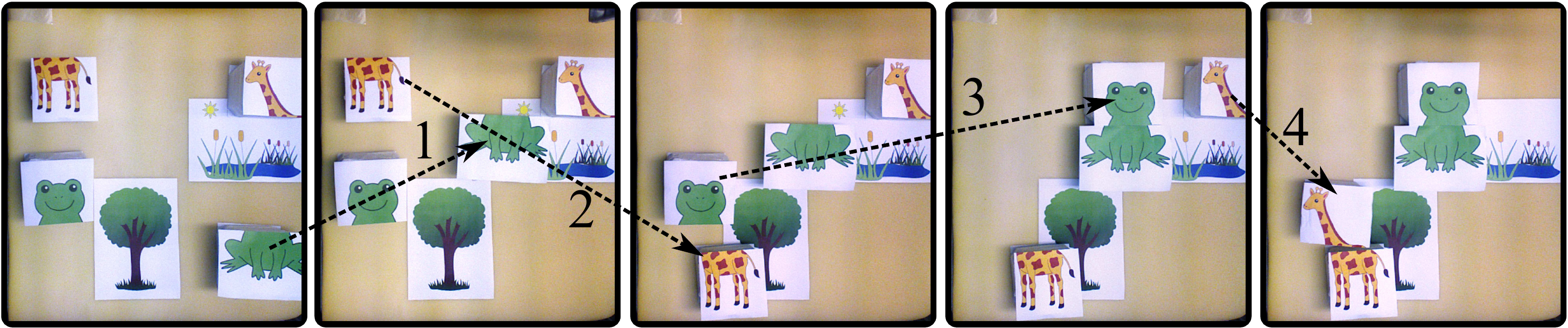} 
      \label{fig:Animal_repro}
      }
      \caption{Animal puzzle. The initial and the final objects' configurations in the \emph{world} are different in (a) and (b). Black arrows show the operations.}
\end{figure}

In this task, the \emph{world} includes two sets of objects which complete a `frog' and a `giraffe' puzzle. There are also two labels (i.e. landmarks) in the \emph{world}, a `pond' and a `tree'. The goal is to assemble the set of objects for each animal with respect to the labels according to the demonstration. Figure~\ref{fig:Animal_demo} shows the sequence of operations in the demonstration phase. To show the capability of generalization, the `tree' and the `pond' labels are randomly replaced by the tutor before the reproduction phase. Figure~\ref{fig:Animal_repro} shows the sequence of operations reproduced by VSL after learning the spatial relationships among objects. This experiment shows the VSL's capability of relative positioning reported in Table~\ref{Table1}. In the previous task, the final objects' configuration in the reproduction and the demonstration phases are always the same. In this experiment, however, by removing the fixed baseline from the \emph{world}, the final result can be a totally new configuration of objects. The accompanying video shows the execution of this task~\citep{iros2013video}.

\subsubsection*{Tower of Hanoi}
\label{subsubsec:task3hanoi}

\begin{figure}[ht]
  \centering
  \includegraphics[width=0.9\columnwidth]{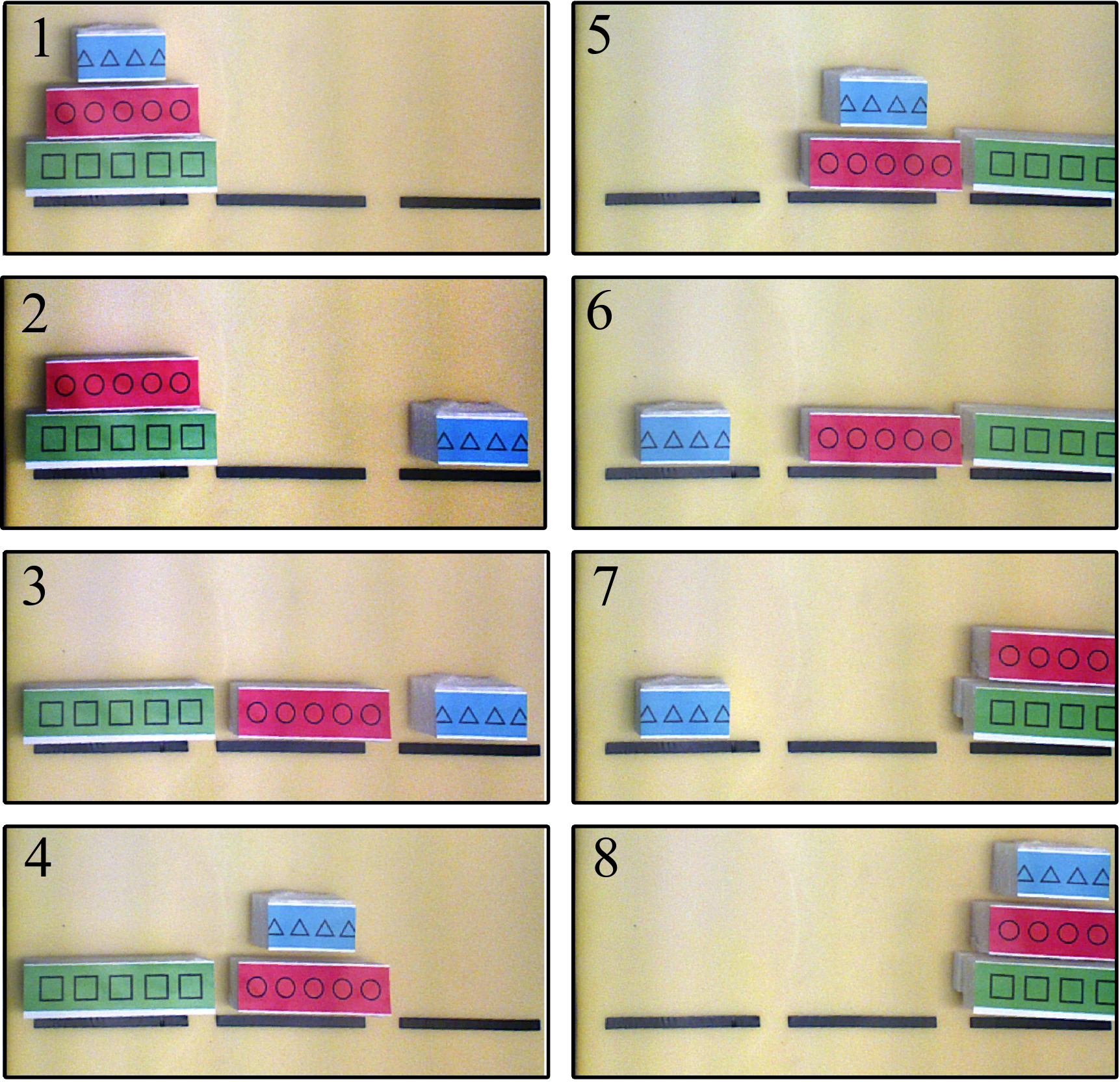} 
  \caption{The sequence of the reproduction for the Tower of Hanoi experiment to illustrate the main capabilities of VSL.}
  \label{fig:Hanoi2}
\end{figure}

In this experiment, the famous Tower of Hanoi puzzle, which consists of a number of disks of different sizes and three bases or rods which actually are landmarks. The objective of the puzzle is to move the entire stack to another rod. This experiment demonstrates almost all capabilities of VSL. Two of these capabilities are not accompanied by the previous experiments. Firstly, our approach enables the user to intervene to modify the reproduction. Such capability can be used to move the disks to another base (e.g. to move the stack of disks to the third base, instead of the second). This can be achieved only if the user performs the very first operation in the reproduction phase and moves the smallest disk on the third base instead of the second. Secondly, VSL enables the user to perform multiple operations on the same object during the learning task. The sequence of reproduction is shown in Figure~\ref{fig:Hanoi2}.

\subsubsection*{Animals vs. Machines: A Classification Task}
\label{subsubsec:Task4animachine}

\begin{figure}[ht]
  \centering
  \includegraphics[width=\columnwidth]{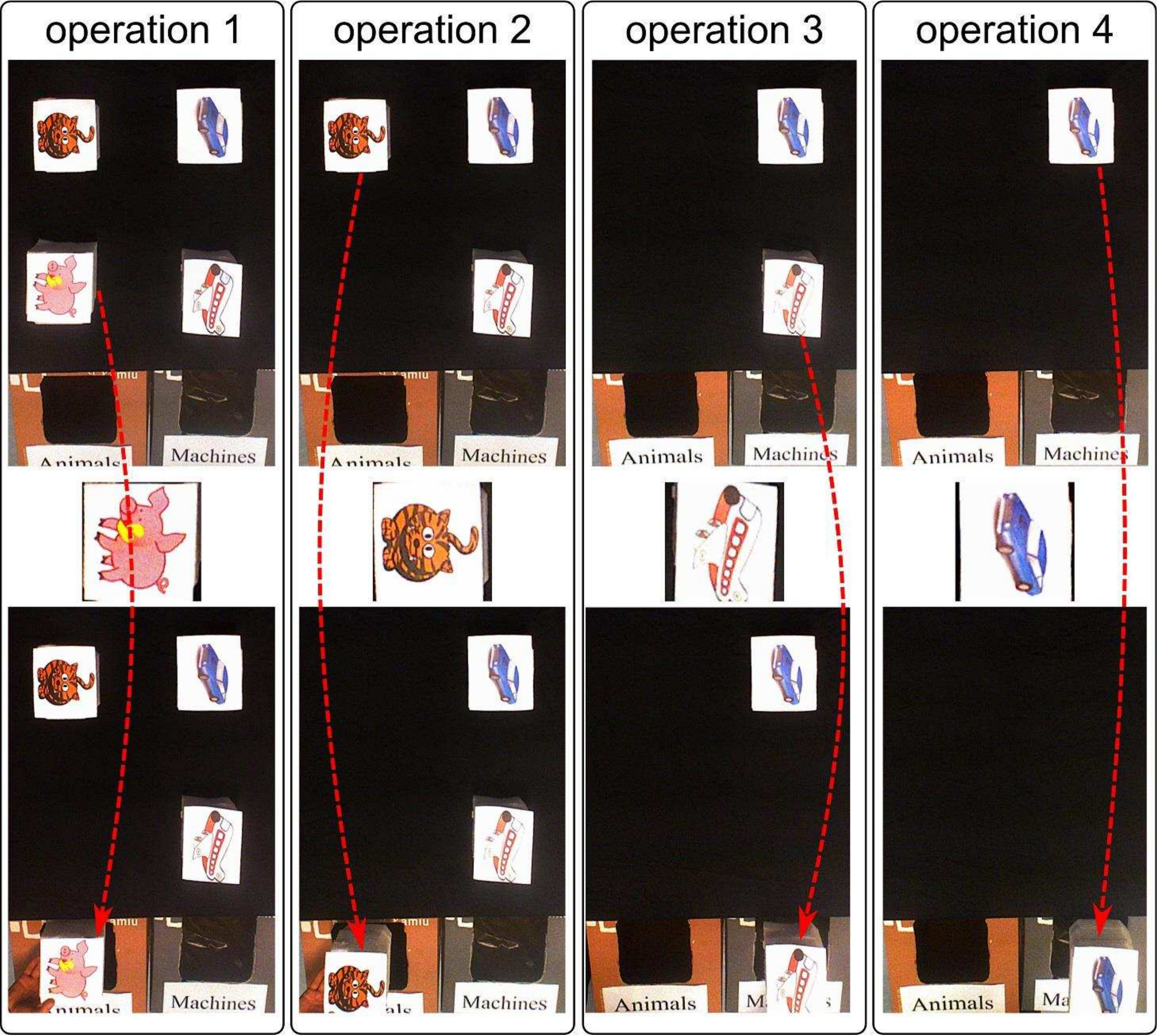} 
  \caption{The sequence of operations in the demonstration phase. Each column represents one pick-and-place operation. In each operation, the tutor picks one object and classifies it either as an `animal' or a `machine'. The selected object in each operation is shown in the middle row.}
  \label{fig:classDemo}
\end{figure}

\begin{figure}[ht]
  \centering
  \includegraphics[width=\columnwidth]{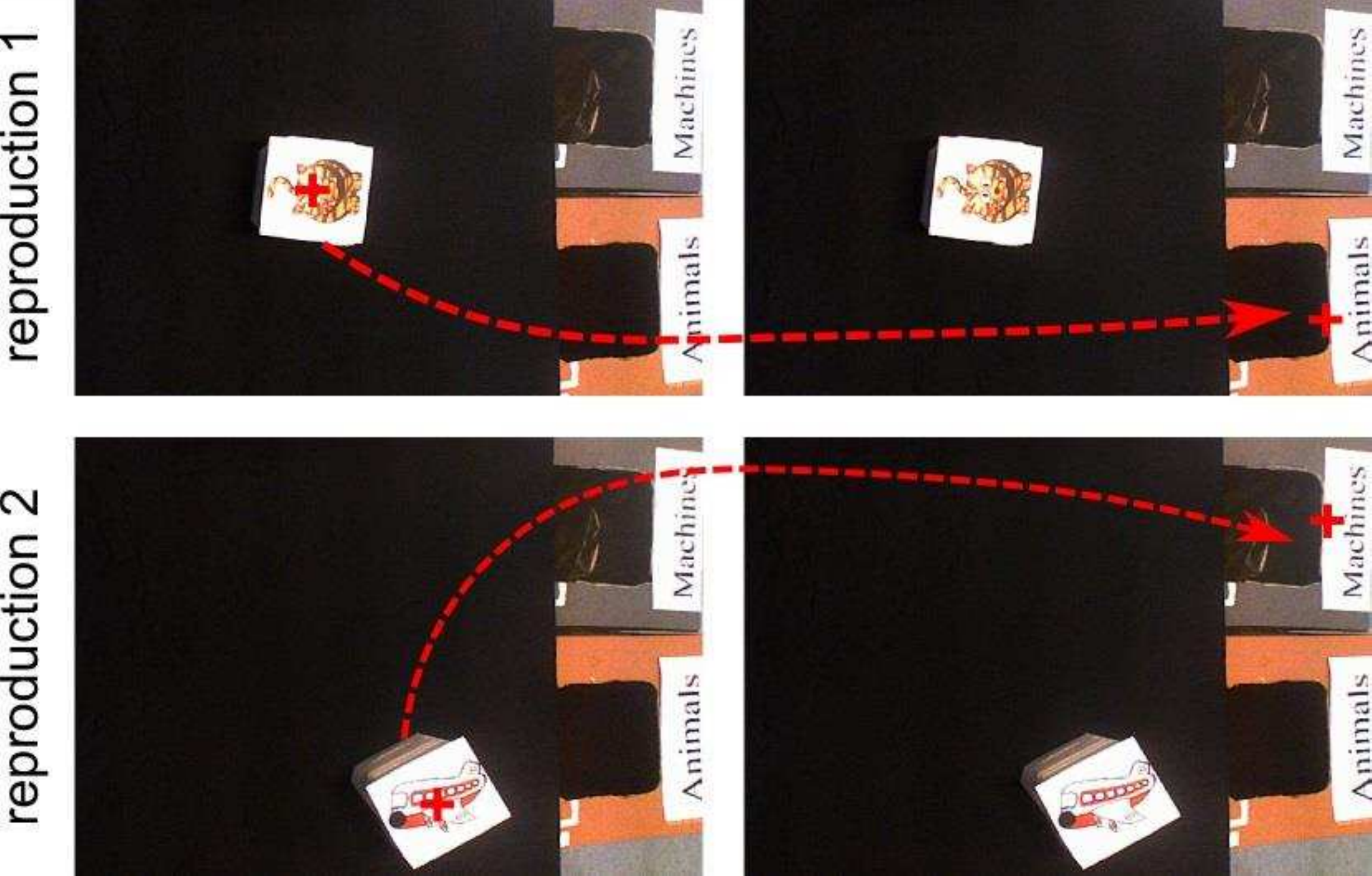} 
  \caption{Two operations during the reproduction phase. Red crosses on objects and on bins show the detected positions for pick and place actions, respectively.}
  \label{fig:classRepro}
\end{figure}

This is an interactive task that demonstrates the VSL capability of classification of objects. In this task, we show that interestingly the same VSL algorithm can be used to learn a classification task. The robot is provided with four objects, two `animals' and two `machines'. Two labeled bins are used in this experiment for classifying the objects. Similar to previous tasks, the objects, labels, and bins are not known to the robot initially. First, all the objects are randomly placed in the \emph{world}. The tutor randomly picks objects one by one and places them in the corresponding bins. Then, in the reproduction phase, the tutor places one of the objects each time, in a different sequence with respect to the demonstration. Differently from previous tasks, the robot does not follow the operations sequentially but searches in the \emph{pre-pick observation} dictionary for the best matching \emph{pre-pick observation}.  The selected \emph{pre-pick observation} is used for reproduction. The tutor can modify the sequence of operations in the reproduction phase by presenting the objects to the robot in a different order with respect to the demonstration. The sequence of operations in the demonstration phase is illustrated in Figure~\ref{fig:classDemo}. Each row represents one pick-and-place operation. During each operation, the tutor picks an object and moves it to the proper bin. The set of \emph{pre-pick} and \emph{pre-place observations} can be seen in left and right columns, respectively. Figure~\ref{fig:classRepro} shows two operations during the reproduction phase. The accompanying video shows the execution of this task~\citep{icar2013video}.

\subsubsection*{Domino: A Turn-taking Task}
\label{subsubsec:task5domino}

\begin{figure*}[ht]
  \centering
  \includegraphics[width=\textwidth]{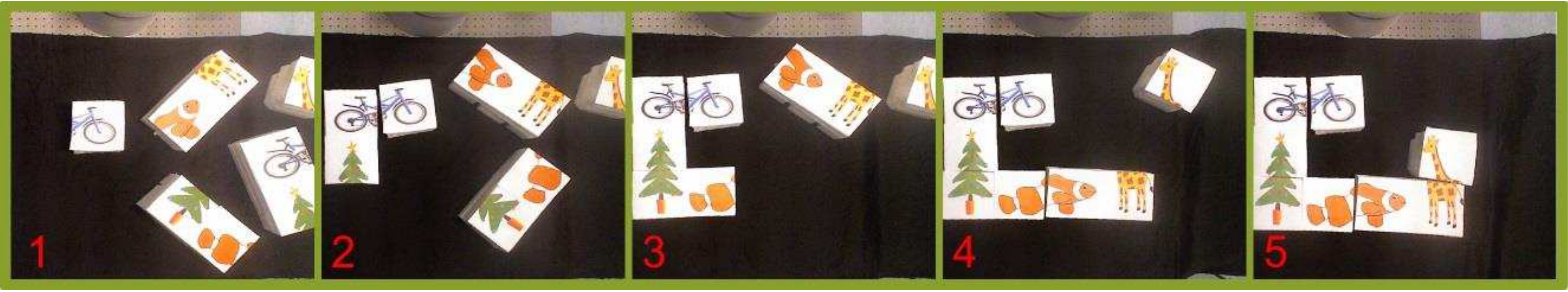} 
  \caption{The sequence of reproduction performed by the robot and the tutor are shown for the turn-taking task of domino. }
  \label{fig:domino}
\end{figure*}

The goal of this experiment is to show that VSL can also deal with the tasks involving the cognitive behavior of turn-taking. In this task, the \emph{world} includes a set of objects all of which are rectangular tiles. Each two pieces of the puzzle fit together to form an object (Figure~\ref{fig:domino}). In the demonstration phase, the tutor first demonstrates all the operations. In order to learn the spatial relationships, the system uses the modified algorithm from the classification task. In the reproduction phase, the tutor starts the game by placing the first object (or another) in a random place. The robot then takes the turn, finds and places the next matching domino piece. The tutor can also modify the sequence of operations in the reproduction phase by presenting the objects to the robot in a different order with respect to the demonstration. The sequence of reproduction is shown in Figure~\ref{fig:domino}.

\section{VSL-3D: An Extension to 3D Tasks}
\label{sec:3dvsl}

In this Section, an extension encompassing VSL is discussed. Although the main structure of the algorithm remains unaltered, VSL-3D requires more sophisticated image processing methods including 3D feature estimation, calculation of surface properties, dealing with noise, 3D match finding, and pose estimation.

\subsection{Point Cloud Processing}
\label{subsec:pointcloud}

\begin{figure}[ht]
    \centering
    \includegraphics[trim=0cm 0cm 0cm 0cm, width=0.9\columnwidth]{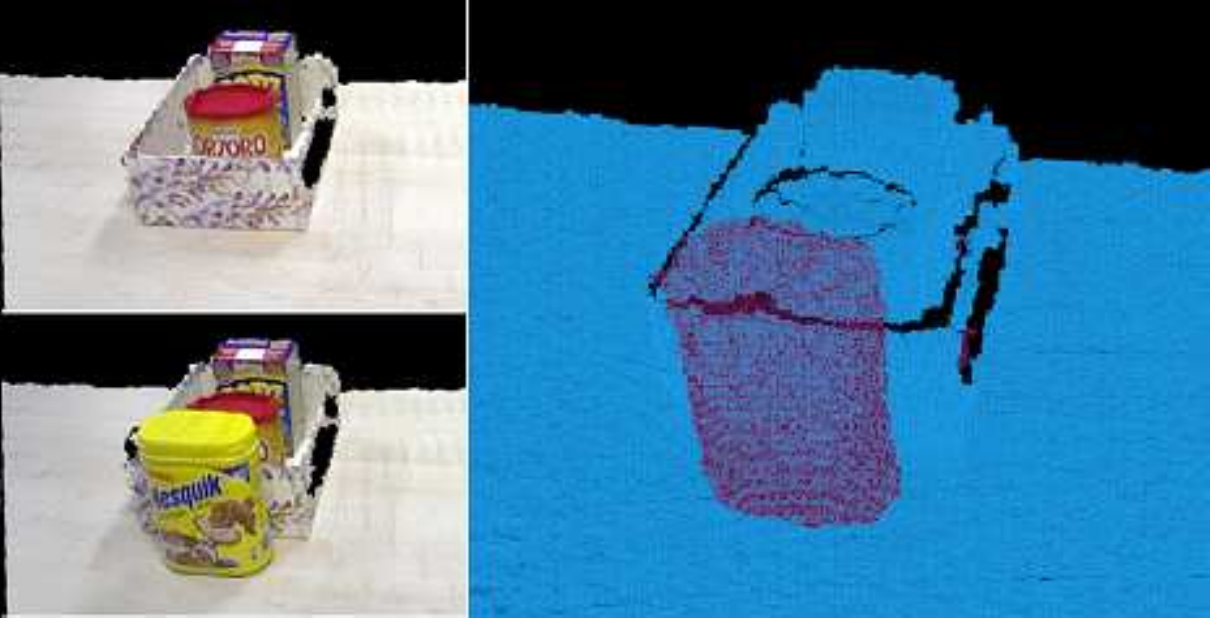}
    \caption{Result of a subtraction process between a pre-pick (top-left) and a pre-place (bottom-left) point clouds. The original clouds are voxelized at a resolution of $2$~mm and the distance threshold for the KNN-search is set to $1$~cm.}
    \label{fig:subtract}%
\end{figure}

For a VSL-3D task, point cloud processing methods should be used in both demonstration and reproduction phases. In the demonstration phase, for each operation, VSL-3D captures one \emph{pre-action} and one \emph{post-action} clouds. The captured raw point clouds are filtered using a pass-through filter to cut the values outside the workspace. This action reduces the number of points by removing unnecessary information (i.e. distant points) from the raw point clouds. The limits of filtering for each axis are derived from the workspace size. From each pair of filtered point clouds, \emph{pre-action} and \emph{post-action observations} are generated. In this state, to increase the speed of further processing, point clouds are downsampled through voxelizing. The points in a voxel are approximated with their centroid. To avoid decreasing the data resolution, the leaf size of the voxel grid must be selected appropriately. The points generated in voxelization phase are stored in a \textit{Kd}-tree structure. Then, $K$ Nearest-Neighbor search (KNN) is applied to extract the downsampled point cloud by finding correspondence between groups of points. The search precision (i.e. error bounds) and the distance threshold should be set properly for the search algorithm. The result of applying the described process on a pair of captured \emph{observations} is depicted in Figure~\ref{fig:subtract}.

\begin{figure*}[ht]
        \centering
        \subfloat[RGB-D]
        {
            \includegraphics[width=0.22\textwidth]{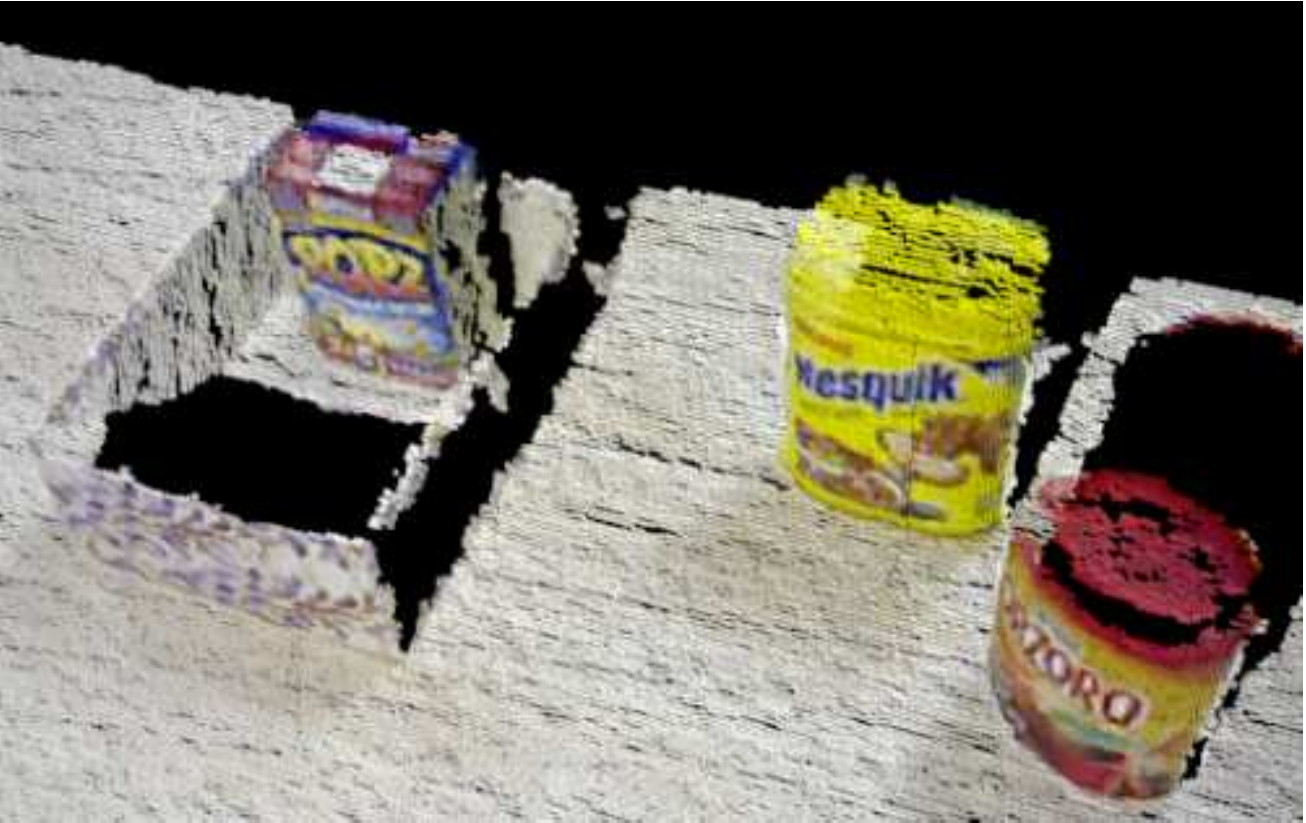}
            \label{fig:rgbd2}
        }
        \subfloat[ICP]
        {
            \includegraphics[width=0.22\textwidth]{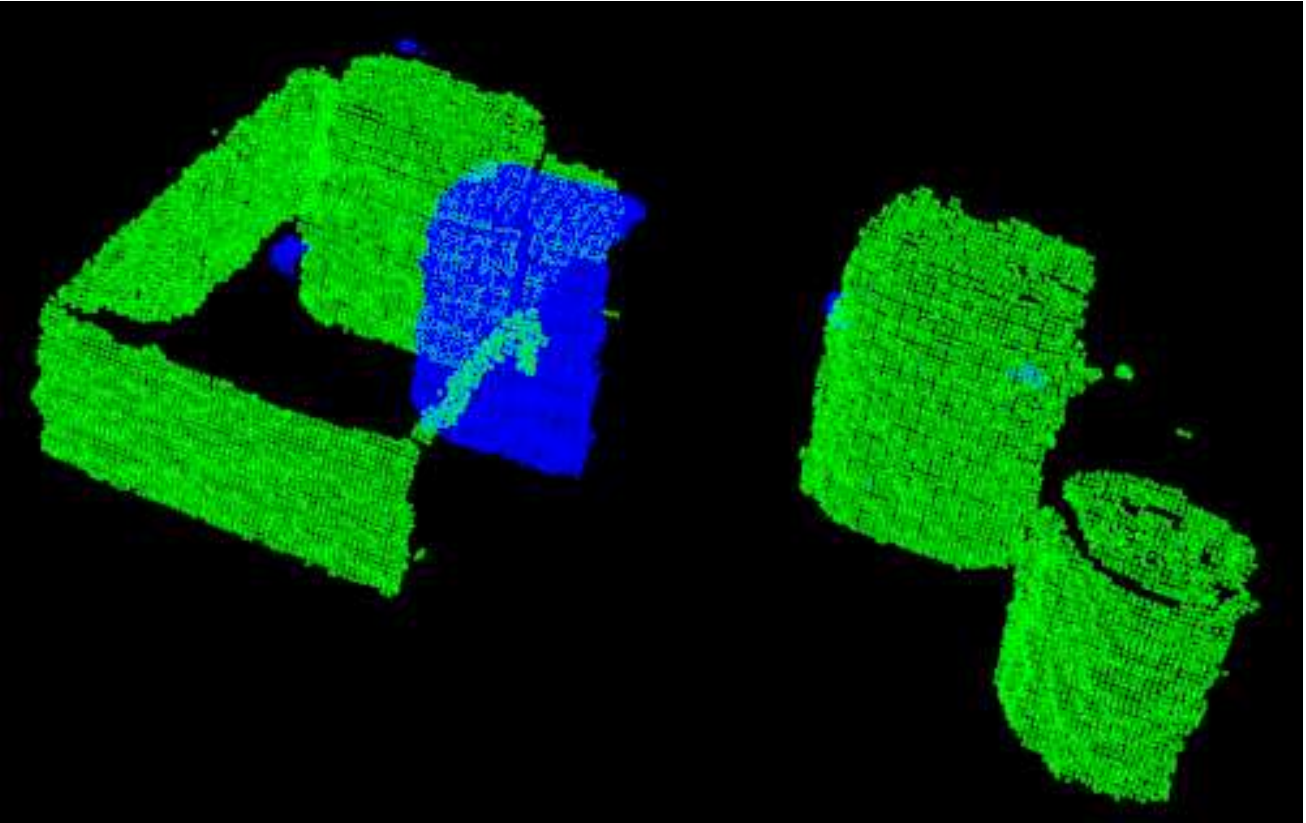}
            \label{fig:icp}
        }
        \subfloat[\tiny{Pre-rejective RANSAC}]
        {
            \includegraphics[width=0.22\textwidth]{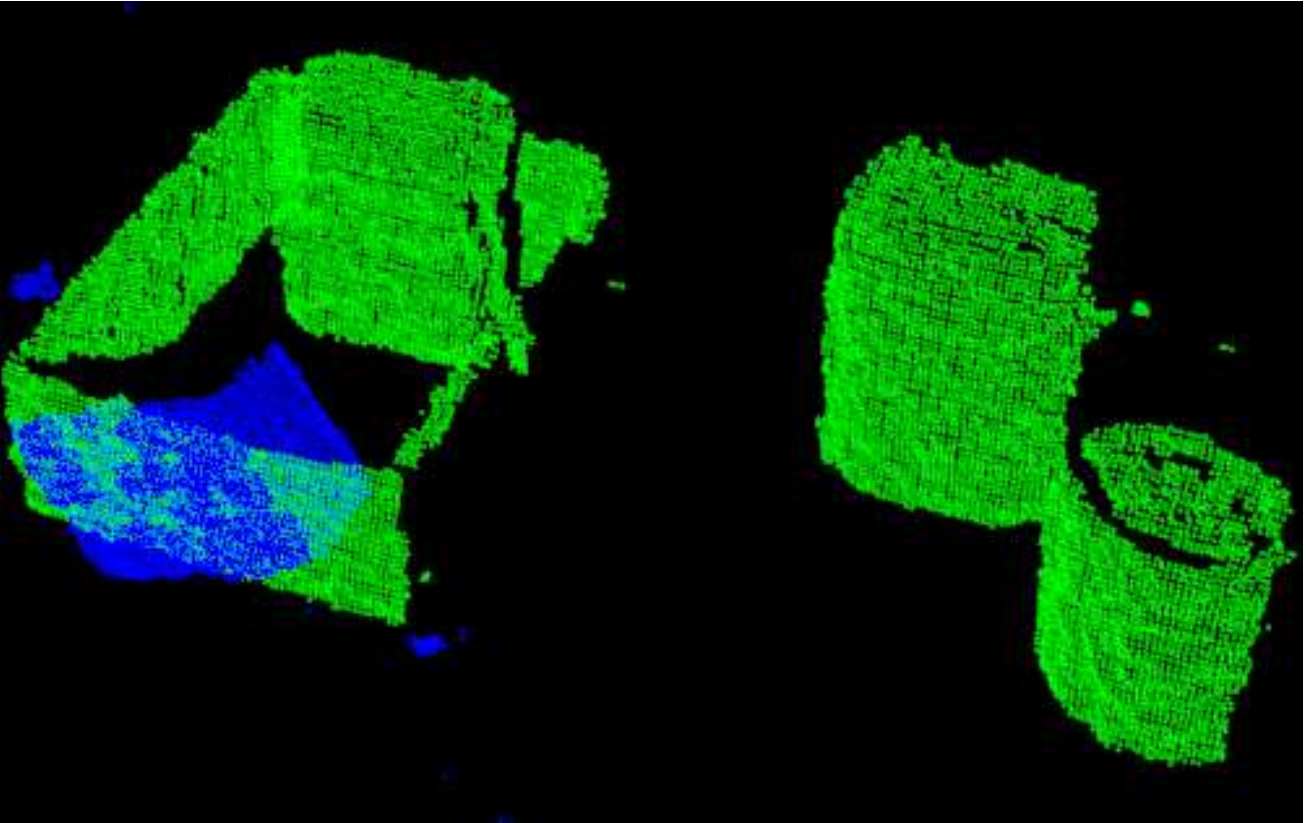}
            \label{fig:ransac}
        }
        \subfloat[SAC-IA]
        {
            \includegraphics[width=0.22\textwidth]{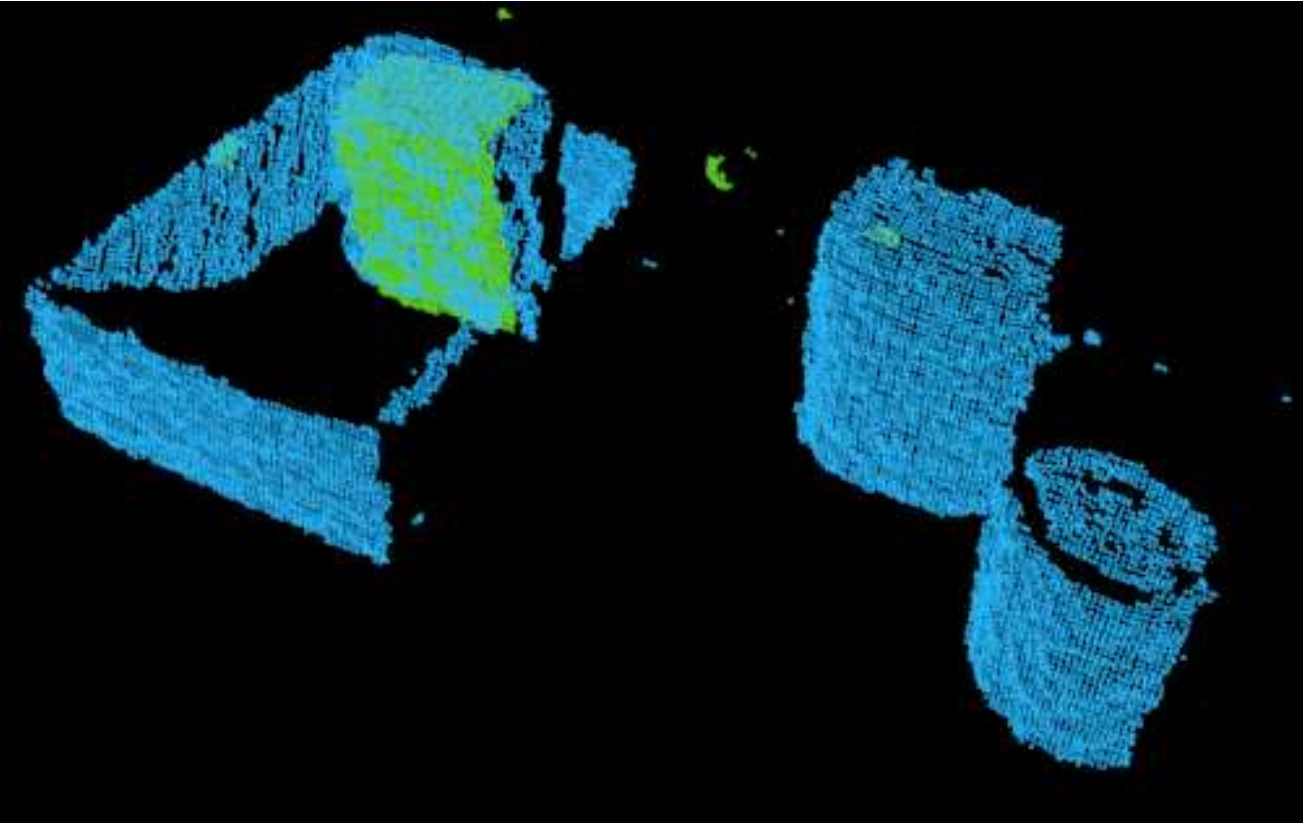}
            \label{fig:sacia}
        }
        \caption{(a) illustrates the captured RGB-D cloud of a scene. The result of the match finding process using different methods are shown in (b), (c), and (d).}
        \label{fig:compare}
\end{figure*}

In the reproduction phase, each captured \emph{world observation} ($\mathcal{W}_{R}$) after being voxelized is compared with the corresponding recorded \emph{observations} in the dictionaries and a metric is applied to find the best match. Surface normals are estimated for each cloud using a \textit{Kd}-tree and search for neighboring points in a pre-defined radius. The estimated surface normals are then used to find local features. There are several methods to represent point features~\citep{steder2010narf}. An acceptable feature descriptor should be able to show the same characteristics in the presence of rigid transformation, different sampling density, and noise. In this section, Fast Point Feature Histogram (FPFH) proposed by~\citet{rusu2009fast} is used which is a computationally efficient version of Point Feature Histogram. FPFH encodes a point's k-neighborhood geometrical properties by generalizing the mean curvature around the point using a multi-dimensional histogram of values. FPFH is chosen because it is 6D pose-invariant and copes with varying sampling densities or noise levels, it depends on the quality of the surface normal estimate at each point. The match finding and pose estimation process for 3D experiments are accomplished by utilizing three different algorithms consisting of Iterative Closest Point (ICP), Pre-rejective RANSAC~\citep{buch2013pose}, and SAmple Consensus Initial Alignment (SAC-IA)~\citep{rusu2009fast}. Figure~\ref{fig:compare} demonstrates some results for the match finding process.

The main differences for implementing VSL in 2D versus 3D are reported in Table~\ref{table:differences}. The table includes the techniques used in this paper. However, the implementation of VSL is not limited to these techniques and depending on the application, various image processing and point cloud processing techniques can be employed.

\begin{table*} [ht]
\caption{The main differences between VSL and VSL-3D approaches.}
\label{table:differences}
  \centering
    \begin{tabular}{|c|c|c|}
        \hline
        & \textbf{VSL}                                                                                                                                  & \textbf{VSL-3D}                                                                                                                                                            \\ \hline
        \small{Observations}   & \small{2D Image}     & \small{3D Point Cloud}    \\ \hline
        \small{Applicable for}  & \small{3DoF Reconfiguration (X, Y, $\theta$)}  & \small{6DoF Reconfiguration} \\ \hline
        \small{Transformation}   & \small{2D to 2D ($3\times3$)}  & \small{3D to 3D ($4\times4$)} \\ \hline
        {\shortstack[c]{ \small{Capturing} \\ \small{observations}}} &
        {\shortstack[c]{ \small{Background Subtraction} \\ \small{Thresholding}}} &
        {\shortstack[c]{ \small{Voxelizing} \\ \small{\textit{Kd}-tree} \\ \small{KNN search }}} \\ \hline
        {\shortstack[c]{\small{Match finding}}}                 & {\shortstack[c]{\small{2D Features}\\ \small{2D Feature Estimation}\\ \small{2D Metrics (SIFT)}\\ \small{RANSAC}}}       & {\shortstack[c]{\small{3D Features}\\ \small{3D Feature Estimation}\\ \small{Voxelizing}\\ \small{Normal/Curvature Estimation}\\ \small{3D Metrics (ICP/RANSAC/SAC-IA)}}}           \\ \hline
        {\shortstack[c]{ \small{Main} \\ \small{assumptions}}} & {\shortstack[c]{\small{No overlap among objects}\\ \small{Objects with the same height} \\ \small{Objects placed at the same level}}} & {\shortstack[c]{\small{Partially visible objects}\\ \small{Objects with different heights}\\ \small{Objects placed in different levels}}} \\ \hline
\end{tabular}
\end{table*}

\subsection{Results}
\label{subsec:experiment2}

In this section, the feasibility and capability of VSL-3D are experimentally validated. The setup is similar to the one described in Section~\ref{subsec:realexp}, where the CCD camera is replaced with an RGB-D sensor namely, Asus Xtion Pro Live.

\subsubsection*{Table Cleaning}
\label{subsubsec:task6Tablecleaning}

\begin{figure*}[ht]
    \centering
    \includegraphics[trim=0cm 0cm 0cm 0cm, width=0.9\textwidth]{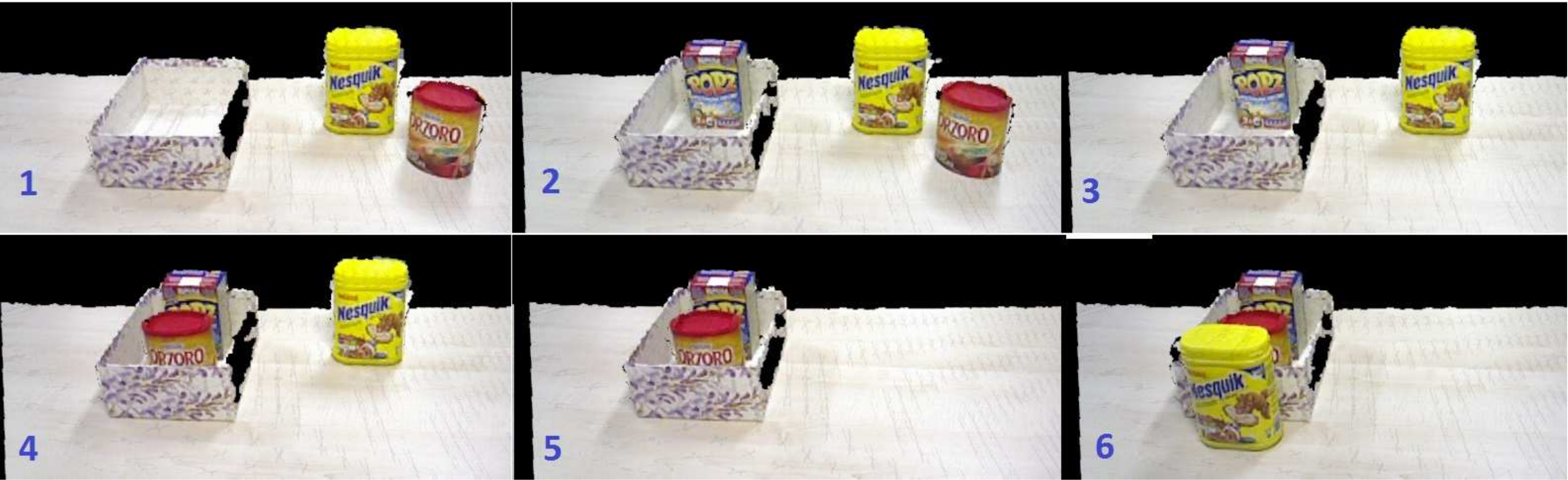}
    \caption{The sequence of operations reproduced by the robot for the table cleaning task. }
    \label{fig:tablecleaning}%
\end{figure*}

In the first experiment with VSL-3D, the \emph{world} includes an empty box and a set of objects which are randomly placed in the workspace. The goal is to clean up the table by picking objects and placing two of them inside the empty box and the final one in front of the box according to the demonstration. The tutor demonstrates the task and the \emph{observations} are captured by the sensor. Before the reproduction phase starts, the objects are reshuffled randomly in the workspace. Figure~\ref{fig:tablecleaning} shows the sequence of operations reproduced by the robot. Although in some operations the objects are occluded by the front wall of the box, the robot can accomplish the task successfully.

\subsubsection*{Stacking Objects}
\label{subsubsec:task7Piling}

\begin{figure*}[ht]
    \centering
        \subfloat[Demonstration]
        {
            \includegraphics[width=0.8\textwidth]{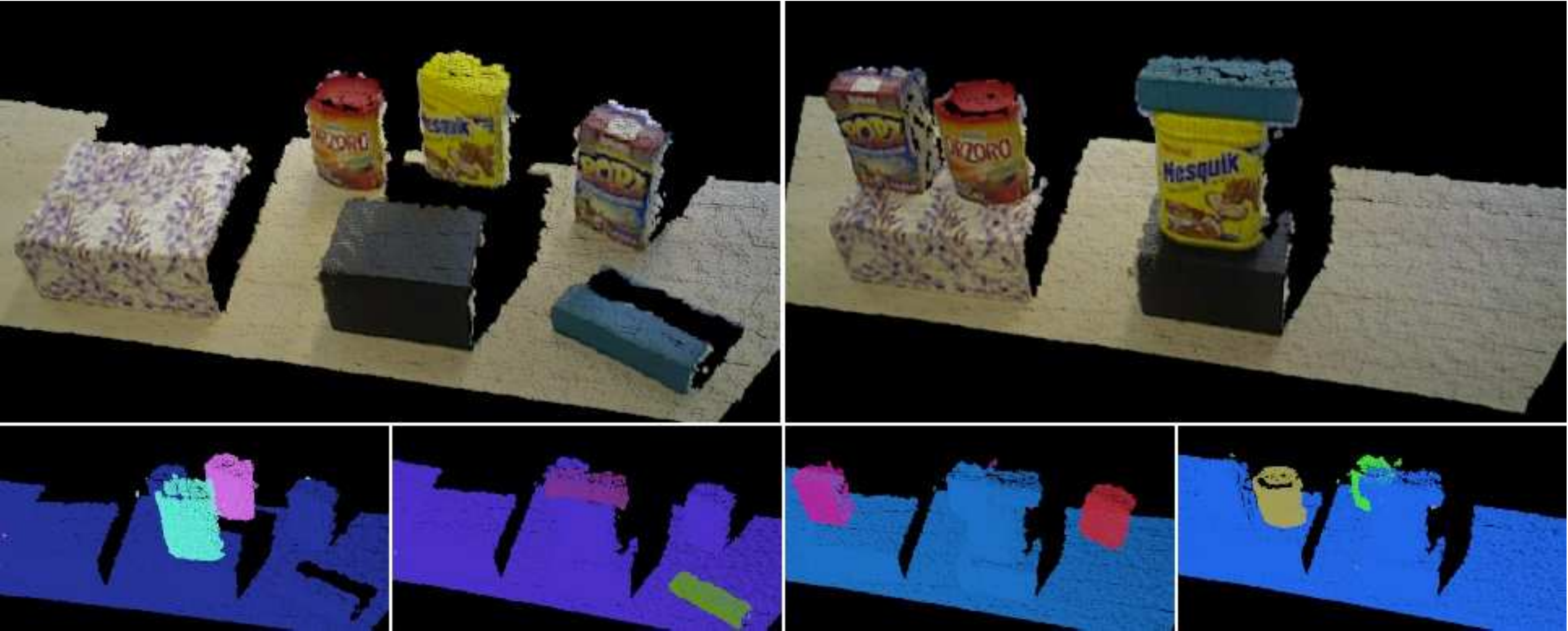}
            \label{fig:pilingDemo}%
        }
        \\
        \subfloat[Reproduction]
        {
            \includegraphics[trim=0cm 0cm 0cm 0cm, width=0.8\textwidth]{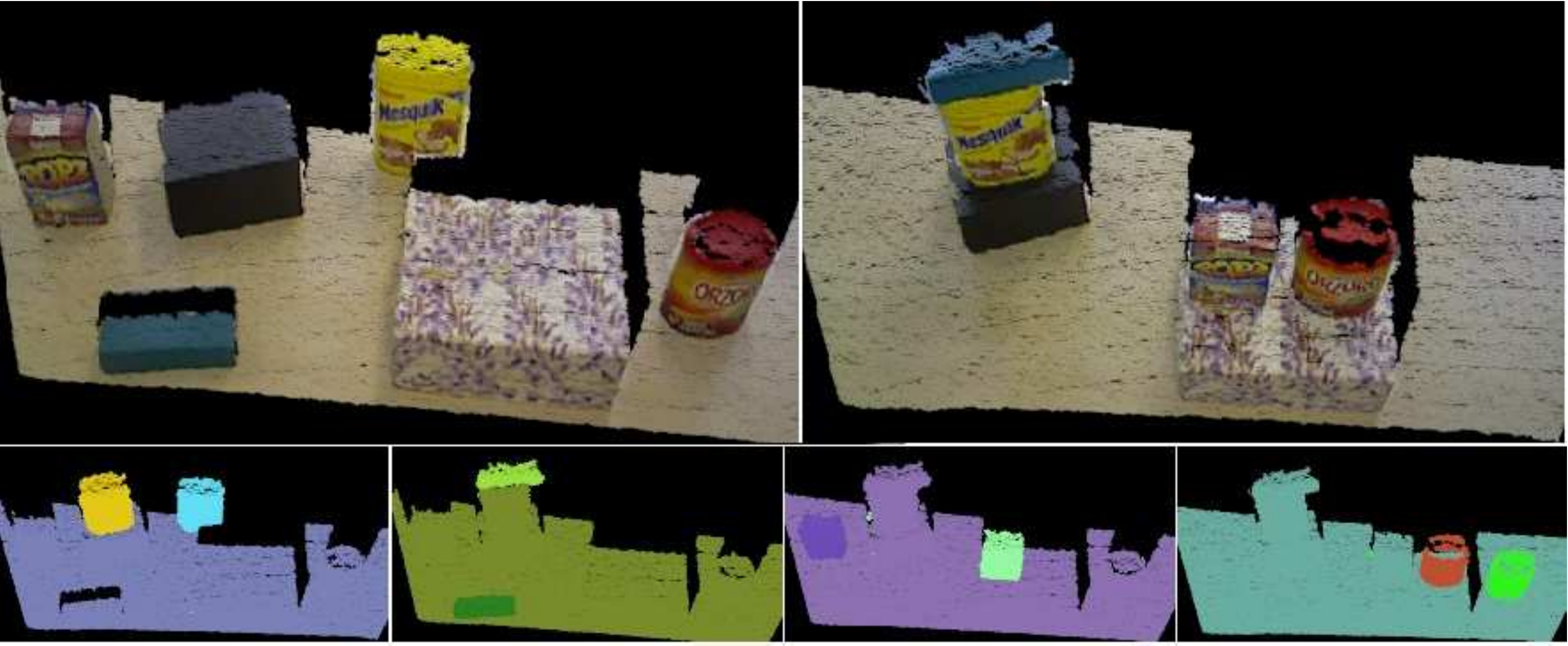}
            \label{fig:pilingRepro}%
        }
        \caption{Demonstration and reproduction phases in a stacking task. The initial and final object configurations are shown in top rows. The bottom rows, from left to right, show the sequence of operations.}
        \label{fig:piling}
\end{figure*}

In the next pick-and-place task, objects are randomly placed on the table (the \emph{world}). The scene includes two dynamic landmarks (i.e. two boxes) which are repositioned by the tutor before starting the reproduction phase. In the first phase, the tutor starts stacking the objects on top of the two boxes. The sequence of operations is captured and together with the initial and final object configurations are shown in Figure~\ref{fig:pilingDemo}. After the learning phase is finished, the robot can reproduce the stacking task in the workspace even if the objects and the boxes are randomly placed. The corresponding sequence of operations is shown in Figure~\ref{fig:pilingRepro}.

\subsubsection*{Results}  %
\label{subsubsec:3dresult}

In both experiments, the \emph{pre-} and \emph{post-action observations} are extracted from a single demonstration. These \emph{observations} are point cloud data which are actually 2.5D and are acquired from a single point of view. Since VSL does not use \emph{a priori} knowledge about objects, if reconfigured in the \emph{world}, they might not be detected and recognized successfully during the reproduction phase. If an object is observed in the demonstration phase, there is no guarantee that it can be detected in the reproduction phase while observed from a new point of view. This is due to object transformations before the reproduction phase. Thus, in order to implement a robust version of VSL-3D, more robust features for both textured and textureless objects and modifications in the point cloud processing part of the algorithm are needed. However, the main steps of the algorithm remain unaltered. Since image processing and point cloud processing are not the focus of this paper, we direct the readers towards appropriate references, e.g. \citep{ekvall2008robot,papazov2012rigid,guadarrama2013grounding}.

\section{Learning Symbolic Representations of Actions using VSL}
\label{sec:symbolic}

As described in Section~\ref{sec:vsl} and Section~\ref{sec:3dvsl}, VSL and VSL-3D allow a robot to identify the spatial relationships among objects and learn a sequence of actions for achieving the goal of a task. We have shown that VSL is capable of learning and generalizing different skills such as object reconfiguration, classification, and turn-taking interactions from a single demonstration. One of the shortcomings of VSL is that the primitive actions such as pick, place, and reach, have to be manually programmed in the robot. For instance, in Section~\ref{subsec:method}, a simple trajectory generation module was devised that is able to produce trajectories for pick and place primitive actions. This issue can be addressed by combining VSL with a trajectory-based learning approach such as Imitation Learning~\citep{ijspeert2001trajectory}. The obtained framework can learn the primitive actions of the task using conventional Learning from Demonstration strategies while it learns the sequence of actions, preconditions, and effects of the actions using VSL. We discuss this solution in this section.

Furthermore, in order to use the advantage of standard symbolic planners, the original planner of VSL can be replaced with an already available action-level symbolic planner, for instance, SGPlan 6 by~\citet{chen2004sgplan}. The integration of VSL and a symbolic planner brings significant advantages over using VSL alone. Learning preconditions and effects of the actions in VSL and making symbolic plans based on the learned knowledge deals with the long-standing important problem of Artificial Intelligence, namely Symbol Grounding~\citep{harnad1990symbol}.

There are a few approaches in which a high-level symbolic representation is formed using low-level action descriptions. The method proposed by~\citet{jetchev2013learning} finds relational, probabilistic STRIPS operators by searching for a symbol grounding that maximizes a metric balancing transition predictability with model size. \citet{konidaris2014constructing} avoid this search by directly deriving the necessary symbol grounding from the actions.

In this Section, a new robot learning framework is proposed by integrating a sensorimotor learning framework (Imitation Learning), VSL, and a symbolic-level action representation and planning layer. The capabilities and performance of the proposed framework, which is called VSL-SP, are validated using real-world experiments with an iCub robot.

\begin{figure}[ht]
    \centering
    \includegraphics[trim=0cm 0cm 0cm 0cm, width=0.9\columnwidth]{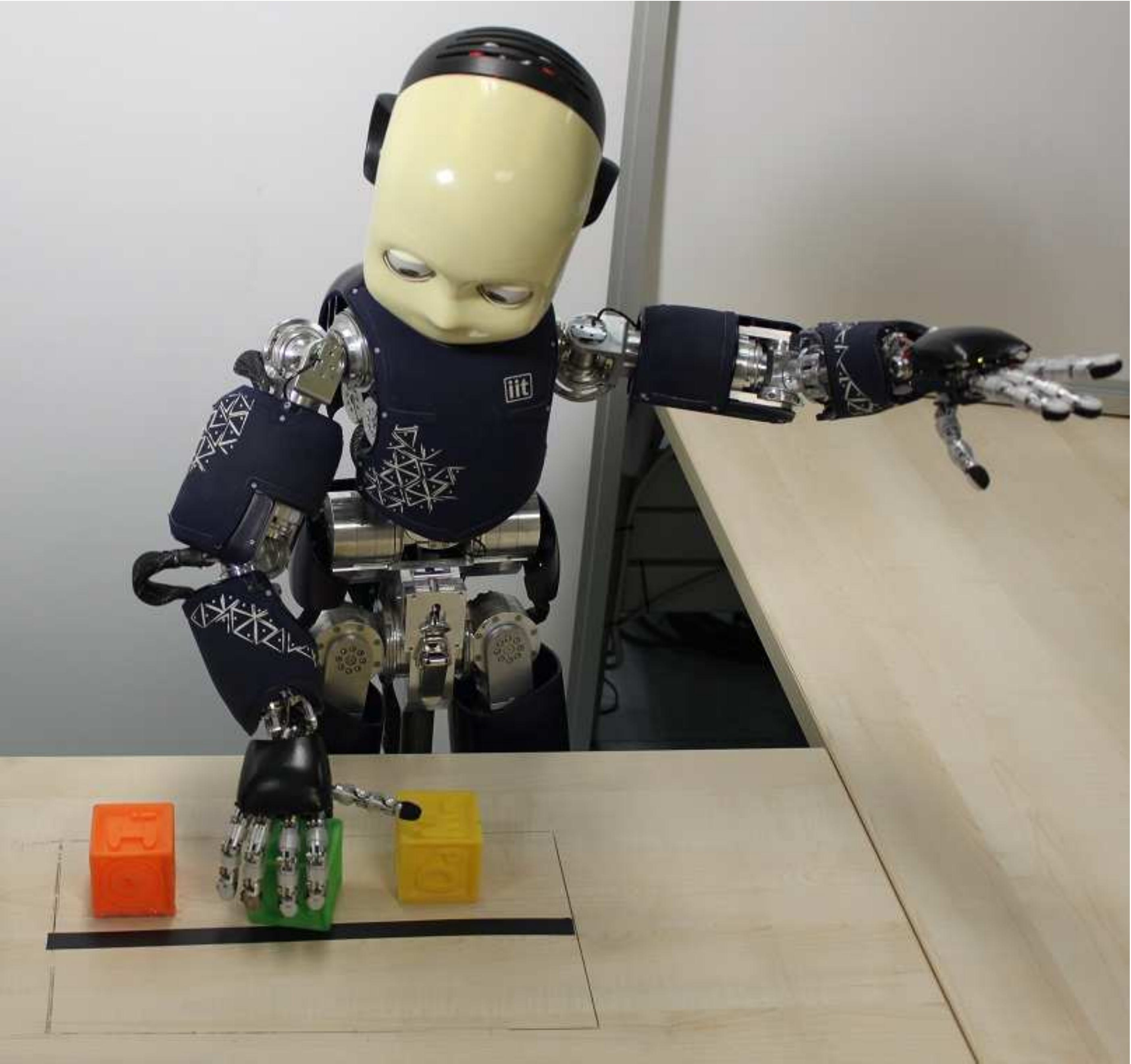}
    \caption{The iCub robot interacting with objects.}%
    \label{fig:icubfirst}%
\end{figure}

\subsection{Overview}
\label{subsec:overview}
The proposed learning approach, VSL-SP, consists of three layers:

\begin{itemize}
\setlength\itemsep{0em}
  \item \textbf{Imitation Learning (IL)}, which is suitable for teaching trajectory-based skills (i.e. actions) to the robot, as it has been posited by~\cite{argall2009survey}. IL relaxes the assumption about having to encode actions manually.
  \item \textbf{Visuospatial Skill Learning (VSL)}, which allows us to capture spatial relationships among an object and its surrounding objects and to extract a sequence of actions to reach the goal of the task. 
  \item \textbf{Symbolic Planning (SP)}, which uses a symbolic representation of actions, to plan or replan the execution of the task. A symbolic planner allows the system to generalize not only to different initial states, but also to different final states.
\end{itemize}

\begin{figure*}[ht]
    \centering
    \includegraphics[trim=0cm 0cm 0cm 0cm, width=0.98\textwidth]{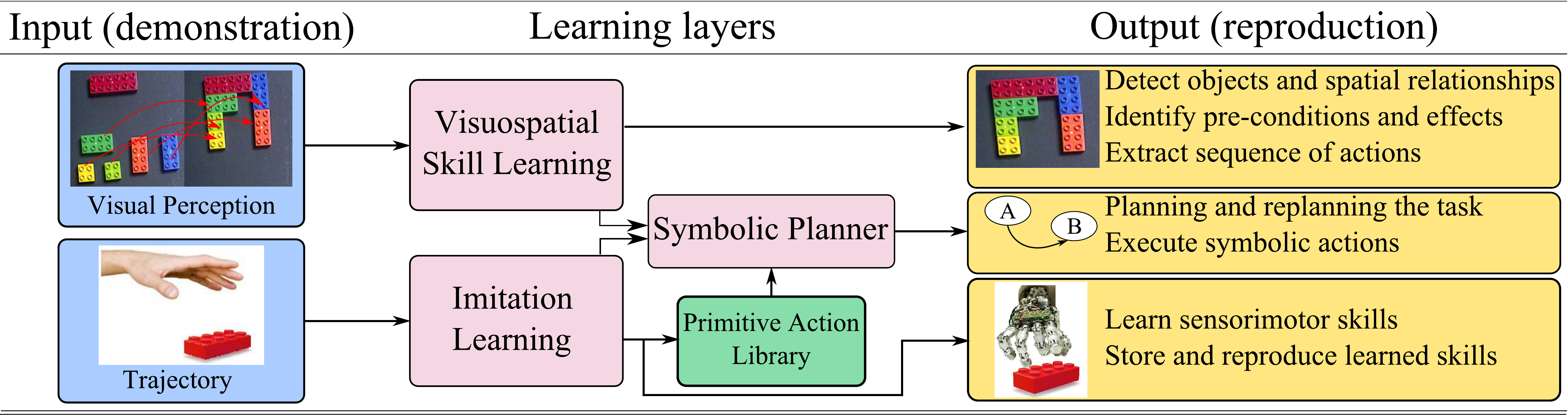}%
    \caption{A flow diagram illustrating the proposed integrated approach, VSL-SP, comprising three main layers. The robot learns to reproduce primitive actions through Imitation Learning. It also learns the sequence of actions and identifies the constraints of the task using VSL. The symbolic planner is employed to solve new symbolic tasks and execute the plans.}
    \label{fig:flowMethod}%
\end{figure*}

A high-level flow diagram of VSL-SP including the three layers together with their input and output can be seen in Figure~\ref{fig:flowMethod}. IL is employed to teach sensorimotor skills to the robot (e.g. pull and push). Learned skills are stored in a primitive action library, which is accessible by the planner. VSL captures an object's context for each demonstrated action. This context is the basis of the visuospatial representation and encodes implicitly the relative position of the object with respect to other objects simultaneously. In order to employ the advantages of symbolic-level action planners, a PDDL 3.0 compliant planner is integrated with IL and VSL to provide a general purpose representation of a planning domain. The identified preconditions and effects modify their formal PDDL definition (i.e. the planning domain). Once formal action definitions are obtained, it is possible to exploit the planning domain to reason upon scenarios that are not strictly related to what has been learned, e.g. including different object configurations, a varied number of objects or different objects at all. One of the main advantages of VSL-SP is that it extracts and utilizes multi-modal information from demonstrations including both the sensorimotor information and visual perception, which is used to build symbolic representation structures for actions.

\subsection{Learning Sensorimotor Skills using Imitation Learning}
\label{subsec:imitationLearning}

IL enables robots to learn and reproduce trajectory-based skills from a set of demonstrations through kinesthetic teaching~\citep{ijspeert2013dynamical}. In VSL-SP, IL is utilized to teach primitive actions (i.e. pull and push) to the robot. In particular, Dynamic Movement Primitives (DMP), which are designed for modeling attractor behaviors of autonomous nonlinear dynamical systems, are utilized~\citep{ijspeert2013dynamical}.

\begin{figure*}[ht]
    \centering
    \includegraphics[trim=0cm 0cm 0cm 0cm, width=\textwidth]{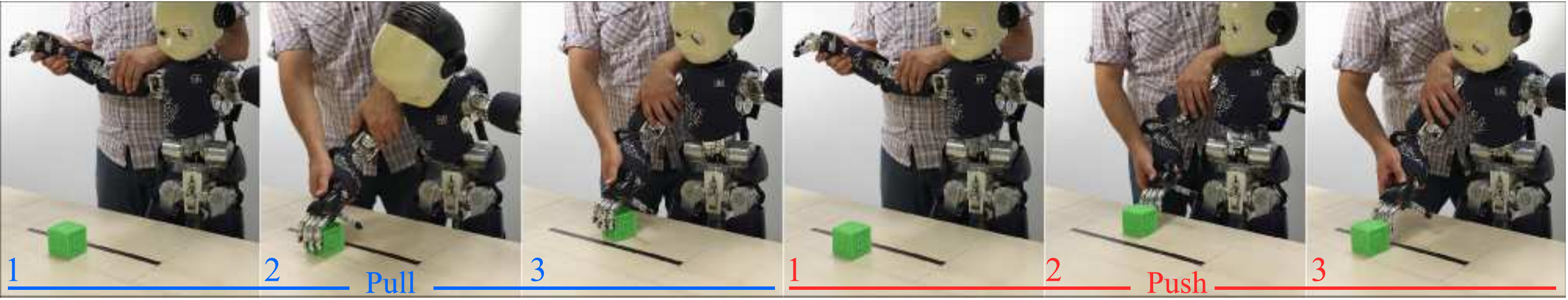}
    \caption{The tutor is teaching the primitive actions including pull and push to the robot using kinesthetic teaching.}
    \label{fig:KinestheticTeaching}%
\end{figure*}

In order to create a pattern for each action, multiple desired trajectories in terms of position, velocity and acceleration have to be demonstrated and recorded by a human operator, in the form of a vector $[ y_{demo}(t), \dot{y}_{demo}(t), \ddot{y}_{demo}(t)]$, where $t \in [ 1,...,P]$ and $P$ is the number of datapoints. A controller converts desired position, velocity, and acceleration, $y$, $\dot{y}$, and $\ddot{y}$, into motor commands. DMP employs a damped spring model that can be modulated with nonlinear terms such that it achieves the desired attractor behavior, as follows: %
\begin{equation}%
\begin{aligned}
\tau\dot{z} &= \alpha_{z}( \beta_{z} ( g-y ) - z ) + f + C_f,\\
\tau\dot{y} &= z.
\end{aligned}
\label{equ:transform}%
\end{equation}

In~\eqref{equ:transform}, position and velocity are represented by $y$ and $z$, respectively, $\tau$ is a time constant and $\alpha_z$ and $\beta_z$ are positive constants. With $\beta_z = \alpha_z / 4$,  $y$ monotonically converges toward the goal $g$. $f$ is the forcing term and $C_f$ is an application dependent coupling term. Trajectories are recorded independently of time. Instead, the canonical system is defined as:
\begin{equation}%
\begin{aligned}
\tau\dot{x} &= - \alpha_{x}x + C_c, \\
\end{aligned}
\label{equ:canonical}%
\end{equation}

\noindent where $\alpha_x$ is a constant and $C_c$ is an application dependent coupling term. $x$ is a phase variable, where $x=1$ indicates the start time and $x$ close to zero means that the goal $g$ has been reached. Starting from some arbitrarily chosen initial state $x_0$ such as $x_0=1$ the state $x$ converges monotonically to zero. The forcing term $f$ in~\eqref{equ:transform} is chosen as follows:
\begin{equation}%
\begin{aligned}
f(x) &= \frac{\sum_{i=1}^N \psi_i(x) \omega_i}{\sum_{i=1}^N \psi_i(x)} x(g-y_0),
\end{aligned}
\label{equ:force}%
\end{equation}

\noindent where $\psi_i \; ,i=1,\ldots,N,$ are fixed exponential basis functions, $\psi_i(x)= \exp(-h_i (x - c_i)^2)$, where $\sigma_i$ and $c_i$ are the width and centers of the basis functions, respectively, and $w_i$ are weights. $y_0$ is the initial state. The parameter $g$ is the target coinciding with the end of the movement $g=y_{demo}(t=P)$, $y_0=y_{demo}(t=0)$. The parameter $\tau$ must be adjusted to the duration of the demonstration. The learning process of the parameters $w_i$ is accomplished with a locally weighted regression method, because it is very fast and each kernel learns independently of others~\citep{ijspeert2013dynamical}.

\begin{figure}[ht]
 \centering
  \subfloat[reach-before sub-action.]%
    {
    \includegraphics[width=0.45\columnwidth]{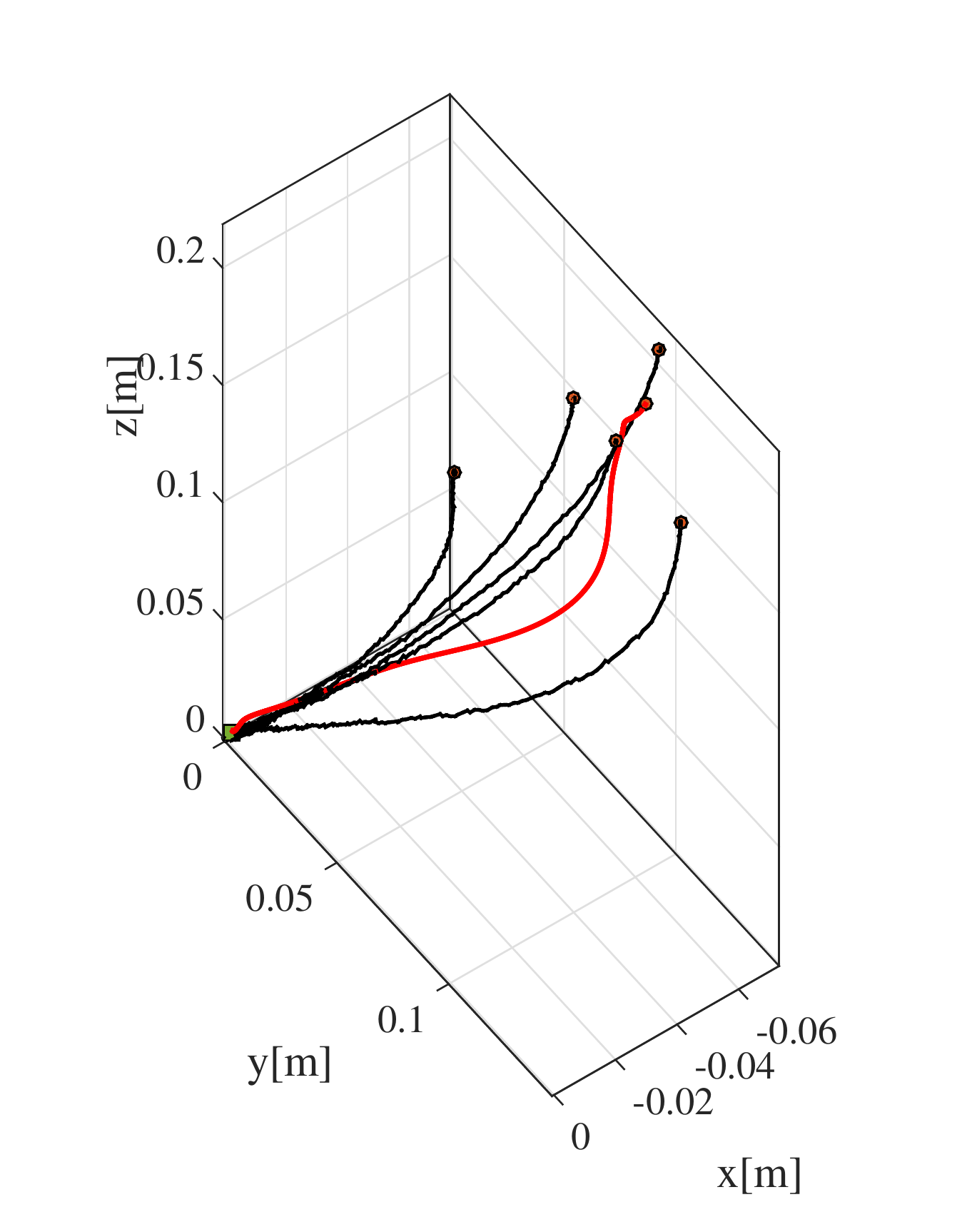}
    \label{fig:reachBefore}
    }
    \hspace{1mm}
  \subfloat[reach-after sub-action.]
    {
    \includegraphics[width=0.45\columnwidth]{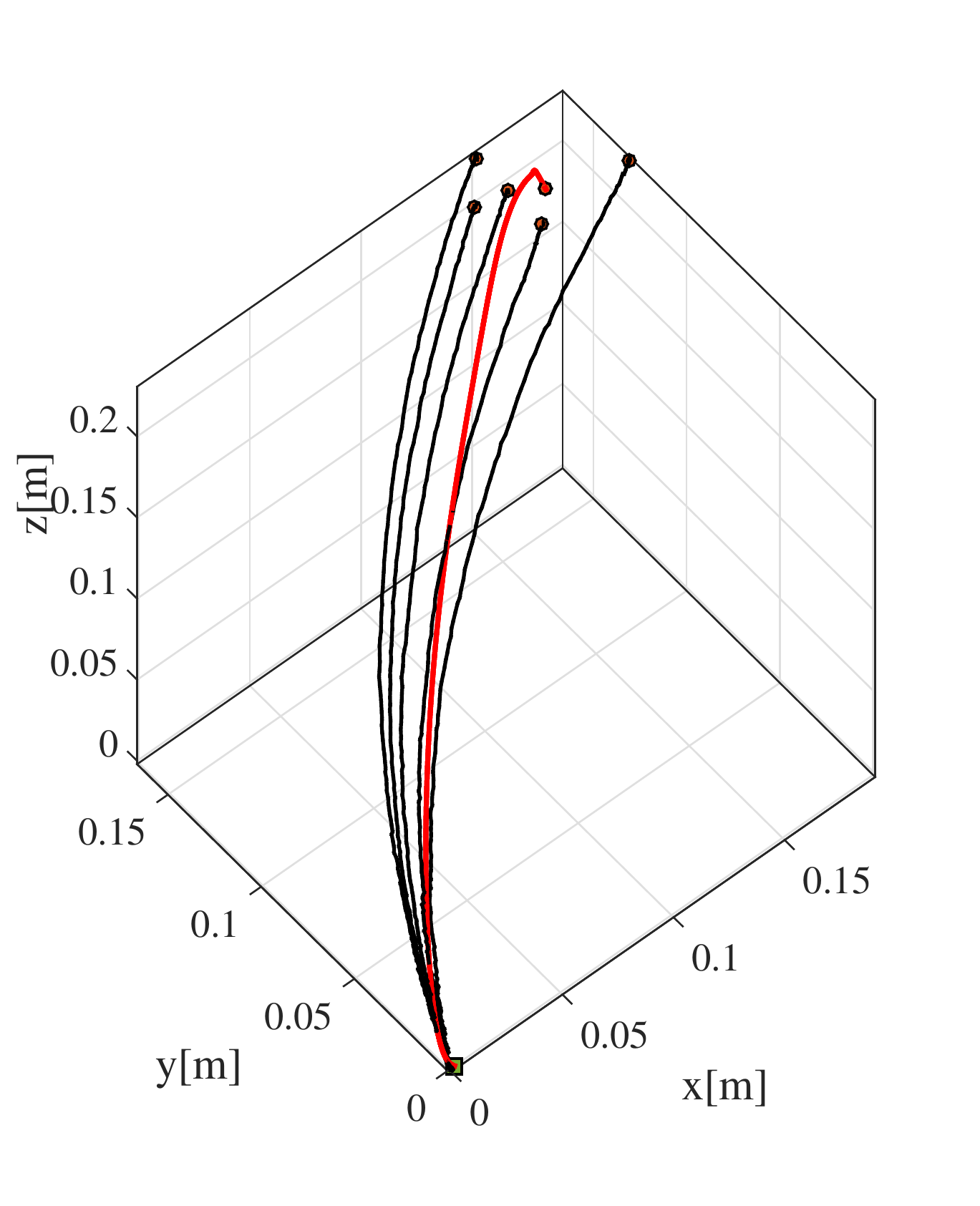}
    \label{fig:reachAfter}
    }
 \caption{The recorded set of demonstrations for two sub-actions are shown in black. The reproduced trajectories from an arbitrary initial position towards the target (the green square) are shown in red.}
\label{fig:imitation_learning}
\end{figure}

In the conducted experiments two primitive actions, namely pull and push are used, each of which is demonstrated to the robot through Kinesthetic teaching and learned using DMP. The robot can reproduce the action from a new initial pose towards the target. To achieve this goal, the \textit{pull} action is decomposed into \textit{reach after} and \textit{move backward} sub-actions and also the \textit{push} action is decomposed into \textit{reach before} and \textit{move forward} sub-actions. The recorded sets of demonstrations for both \textit{reach before} and \textit{reach after} sub-actions are depicted as black curves in Figure~\ref{fig:imitation_learning}. Each red trajectory in Figure~\ref{fig:imitation_learning} illustrates a reproduction. In both Figures, the goal (i.e. the object), is shown in green. Finally, the learned primitive actions are stored in the primitive action library. Later, the planner connects the symbolic actions to the library of actions. The \textit{move forward} and \textit{move backward} sub-actions are straight line trajectories that can be learned or implemented directly into the trajectory generation module.

\subsection{Learning Action Sequences by VSL}
\label{subsec:learnActionsequence}

At this stage, a library of primitive actions has been built. The robot learns the sequence of actions required to achieve the goal of the demonstrated task through VSL. It also identifies spatial relationships among objects by observing demonstrations. As shown in Figure~\ref{fig:flowMethod}, both IL and VSL observe the same demonstrations (not necessarily at the same time) but utilize different types of data. IL uses sensorimotor information (i.e. trajectories), whereas VSL uses visual data.

\subsection{Generalization of Learned Skills as Symbolic Action Models}
\label{subsec:actionGrounding}

In order to use the learned primitive actions for general-purpose task planning, the actions need to be represented as a symbolic, discrete, scenario-independent form. To this aim, a symbolic representation for both the pull and push actions based on the PDDL 3.0 formalism is defined. However, it is necessary to learn preconditions and effects for each action in the primitive action library. The sensorimotor information in the skill learning process is exploited to derive symbolic predicates, instead of defining them manually.

\begin{figure}[ht]
    \centering
    \includegraphics[trim=0cm 0cm 0cm 0cm, width=\columnwidth]{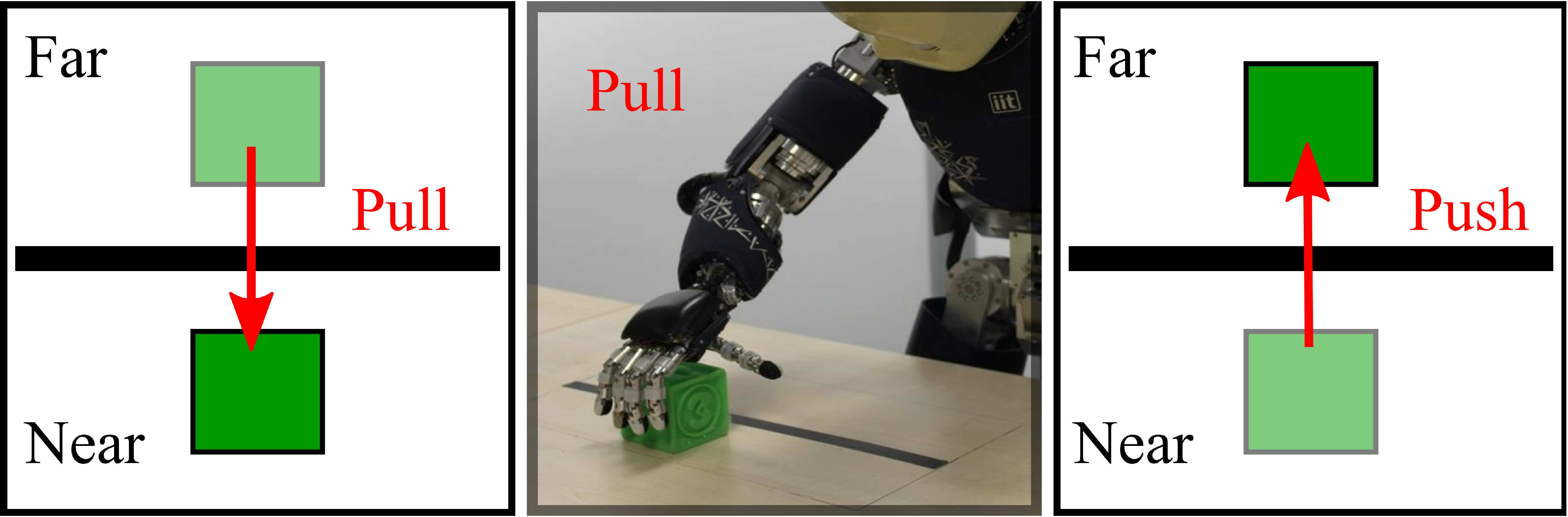}%
    \caption{By extracting the preconditions and effects of the primitive actions while executing them, the robot generalizes the learned sensorimotor skills as symbolic actions.}
    \label{fig:PushPullGrounding}%
\end{figure}

As shown in Figure~\ref{fig:PushPullGrounding}, the robot first perceives the initial workspace in which one object is placed in \texttt{far}. Then the robot is asked to execute the pull action, and after the action is finished, the robot perceives the workspace again. The \emph{pre-action} and \emph{post-action observations} are shown in Figure~\ref{fig:PushPullGrounding}. The robot uses VSL to extract the preconditions and effects of the pull action considering the landmark in the workspace (i.e. the black line in this case). The same steps are repeated for the push action. The domain file in the PDDL formalism is updated automatically by applying this procedure. The preconditions and effects for the pull and push actions are reported in Table~\ref{table:ActionGrounding}.

\begin{table}[h]
    \caption{The preconditions and effects extracted by VSL and written to the planner's domain file.} 
    \label{table:ActionGrounding}
    \centering
    \begin{tabular}{|l|l|l|}
    \hline
    \textbf{Action}        & \multicolumn{1}{c|}{\textbf{push}} & \multicolumn{1}{c|}{\textbf{pull}} \\ \hline
    \textbf{precondition} & \multicolumn{1}{c|}{ \texttt{$\neg$ (far ?b)}}  & \multicolumn{1}{c|}{\texttt{(far ?b)}}              \\ \hline
    \textbf{effect}        & \multicolumn{1}{c|}{\texttt{(far ?b)}}         & \multicolumn{1}{c|}{ \texttt{$\neg$ (far ?b)}}      \\ \hline
\end{tabular}
\end{table}

\subsection{Implementation}
\label{subsec:implementationSymbolic}

In this step, it is possible to generalize each action as well as the demonstrated action sequences, using a PDDL 3.0 compliant formalism. It is noteworthy that in the present work, symbolic knowledge has not been bootstrapped from scratch. On the contrary, starting from the knowledge of the performed action types during the demonstration, symbolic-level knowledge is updated with two kinds of constraints.
The former class includes constraints (in the form of PDDL predicates) related to preconditions and effects. This is done by mapping the elements of observation dictionaries $O^{Pre}$ and $O^{Post}$ to relevant predicates. In the considered scenario, one predicate is enough to characterize push and pull actions, namely \texttt{(far ?b)}, where \texttt{far} is the predicate name and \texttt{?b} is a variable that can be grounded with a specific object that can be pushed and pulled. Specifically, \texttt{push(?b)} is an action that makes the predicate truth value switching from \texttt{$\neg$(far ?b)} to \texttt{(far ?b)}, whereas the opposite holds for \texttt{pull(?b)}. Visual information about the location of objects in the \emph{World} is mapped to the proper truth value of the \texttt{(far ?b)} predicate. The execution of a \texttt{push} demonstration on a green cube (as processed by VSL) allows the system to understand that before the action \texttt{$\neg$(far $B_{green}$)} holds, after the action \texttt{(far $B_{green}$)} holds, and the two predicates contribute to the \texttt{:precondition} and \texttt{:effect} lists of a \texttt{push} action defined in PDDL.

The second class includes constraints (in the form of PDDL predicates) related to the trajectory of the plan that is executed in the demonstration. This can be done analyzing the sequence of actions as represented in the VSL model (i.e. the implicit chain defined by predicates in $O^{Post}$ and $O^{Pre}$ observation dictionaries). In PDDL 3.0, it is possible to define additional constraints, which implicitly limit the search space when the planner attempts to find a solution to the planning problem. In particular, given a specific planning domain (in the form of push and pull actions, as obtained by reasoning on the VSL model), the \texttt{problem} can be constrained to force the planner to follow a specific trajectory in the solution space. One possible constraint is to impose, for instance, the sequence of satisfiability of predicate truth values. As an example, let us assume the VSL model represents the fact that \texttt{$B_{green}$} must be always pushed after \texttt{$B_{blue}$}. According to our formal definition, this implies that the predicate \texttt{(far $B_{green}$)} must hold sometime after the predicate \texttt{(far $B_{blue}$)} holds. To encode this constraint in PDDL formalism, it is sufficient to use the constraint \texttt{(somewhere-after (far $B_{green}$) (far $B_{blue}$))}. If the constraint is used in the planning problem, the planner attempts to find a plan where this sequence is preserved, thereby executing \texttt{push($B_{green}$)} strictly after \texttt{push($B_{green}$)}. Finally, it is noteworthy that such a constraint is removed, the planner is free to choose any suitable sequence of actions to be executed (independently of the demonstration), provided that the goal state is reached.

\subsection{Results}
\label{subsec:experiment3}

\begin{figure}[ht]
    \centering
    \includegraphics[trim=0cm 0cm 0cm 0cm, width=\columnwidth]{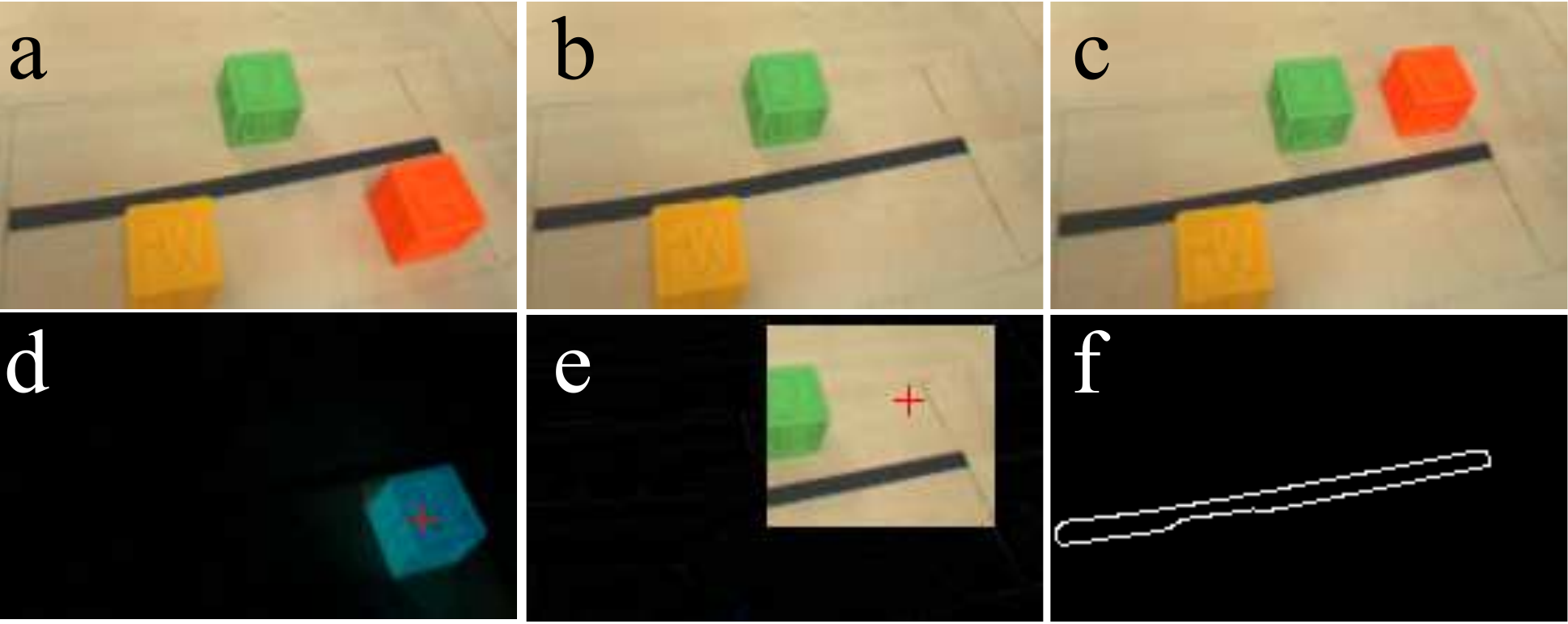}
    \caption{(a) and (b) are rectified pre-action and post-action observations; (c) is a rectified pre-action observation for the next operation; (d) is obtained by background subtraction between (a) and (b); (e) is obtained by background subtraction between (b) and (c); (f) is the result of detecting the border line for spatial reasoning.} 
    \label{fig:imageProcessing}%
\end{figure}

In order to evaluate the capabilities and performance of VSL-SP, a set of experiments are performed with an iCub humanoid robot. As shown in Figure~\ref{fig:icubfirst}, the robot is standing in front of a tabletop including some polyhedral objects. All the objects have the same size but different colors and textures. First, the tutor teaches a set of primitive actions, including pull and push, to the robot using Imitation Learning. For each action, a set of demonstrations is captured through Kinesthetic teaching, while the robot is perceiving the workspace (Figure~\ref{fig:KinestheticTeaching}). Using Imitation Learning, the robot learns to reproduce each action from different initial positions of the hand towards the desired object. The learned primitive actions are stored in a library. In addition, to introduce the concept of \texttt{far} and \texttt{near}, a black landmark, which represents a borderline, is placed in the middle of the robot's workspace. The robot can detect the landmark using conventional edge detection and thresholding techniques. Also, the size of the \emph{frames} ($\mathcal{F}_{D},\mathcal{F}_{R}$) are defined equal to the size of the \emph{world}. Figure~\ref{fig:imageProcessing} shows some instances of image processing results from the conducted demonstrations. In this Section, blob detection and filtering by color techniques are utilized by considering that the iCub robot is capable of detecting the height of the table at the beginning of each experiment by touching it. To detect the object which is moved and the place that the object has been moved to, background subtraction is used. Moreover, a simple (boolean) spatial reasoning has been applied that utilizes the detected pose of the object and the detected borderline and then decides that the object is \texttt{far} or \texttt{near}. However, to extract more sophisticated constraints, qualitative spatial reasoning languages such as Region Connection Calculus by~\citet{randell1992spatial} can be employed.

\subsubsection*{A Simple VSL Task}
\label{subsubsec:taskFirstExp}

\begin{figure*}[ht] 
 \centering
  \subfloat[Demonstration.]%
    {
    \includegraphics[trim=0cm 0cm 0cm 0cm, width=0.65\textwidth]{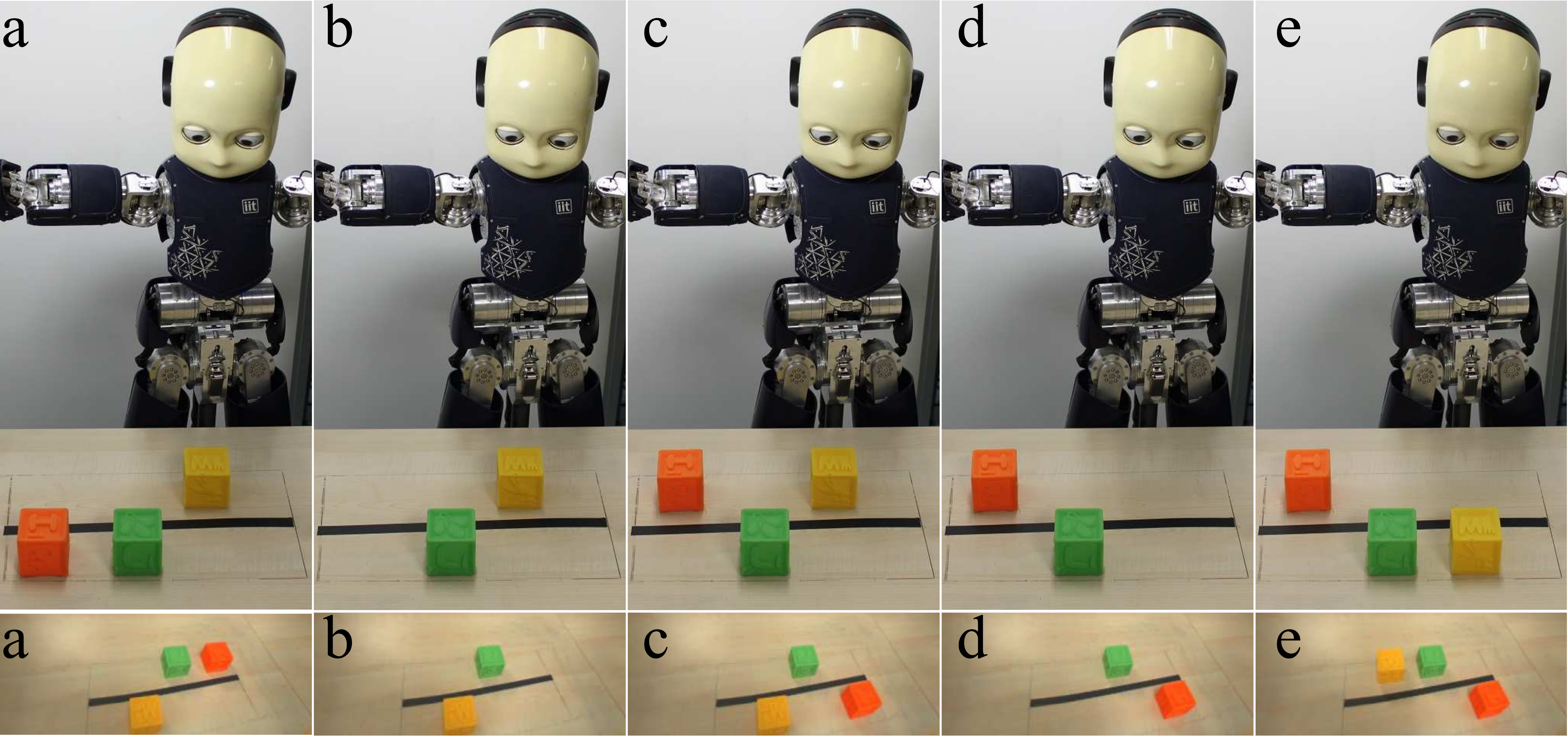}
    \label{fig:demoVSL1}
    }
    \hspace{1mm}
  \subfloat[Reproduction.]
    {
    \includegraphics[trim=0cm 0 0 0,width=0.25\textwidth]{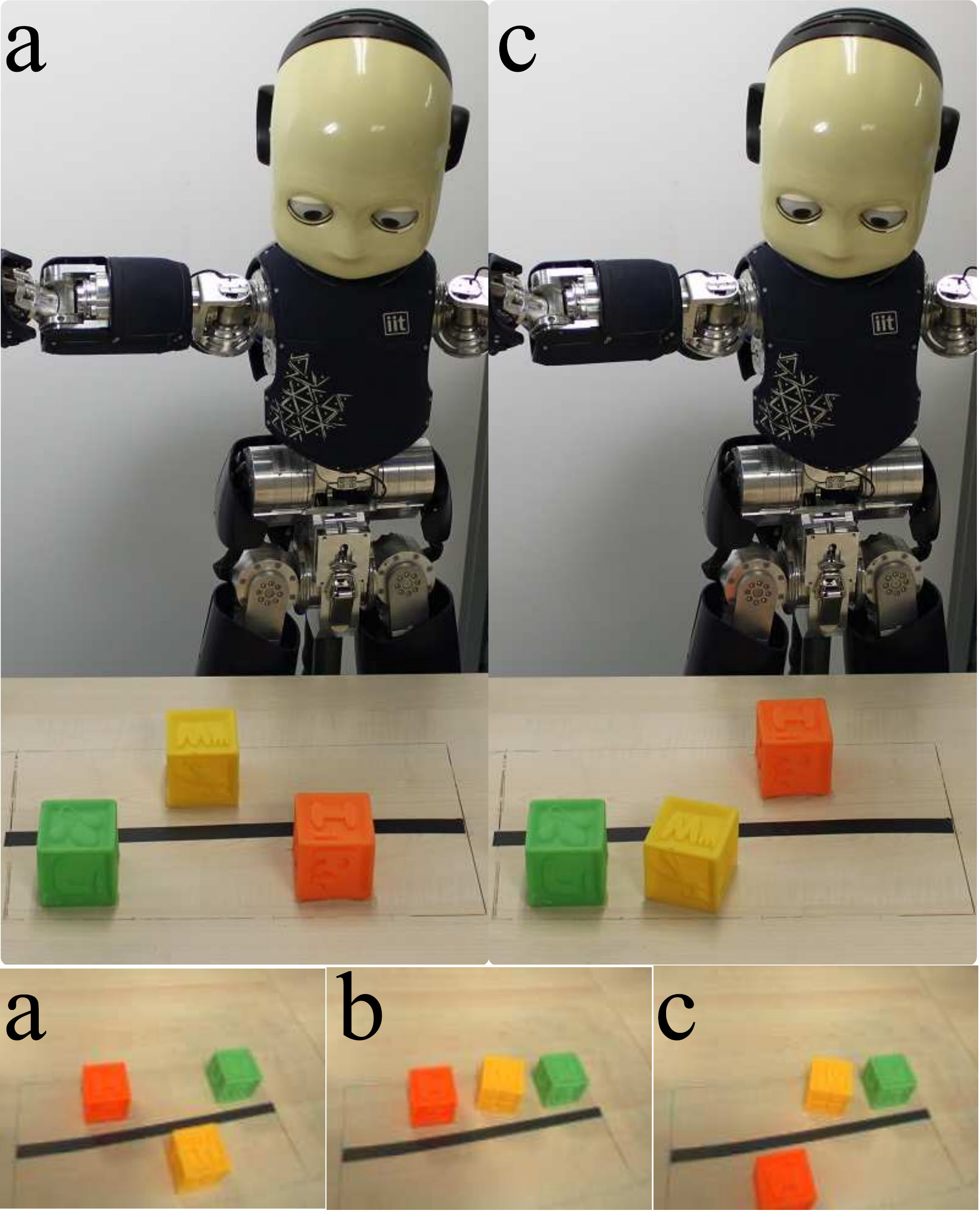}
    \label{fig:reproVSL1}
    }
 \caption{A simple VSL task. In both sub-figures, the bottom rows show robot's point of view.}
\label{fig:traj2}
\end{figure*}

This experiment evaluates the obtained connection between the symbolic actions and their equivalent sensorimotor skills. Initially, there are three objects on the table. The tutor provides the robot with a sequence of actions depicted in Figure~\ref{fig:demoVSL1}. The robot perceives the demonstration, records the observations and extracts the sequence of actions using VSL. In the reproduction phase, objects are reshuffled in the workspace. For the reproduction of the task, instead of SP (Symbolic Planner), the internal planner of VSL is exploited. The result shows that starting from an arbitrary initial configuration of the objects in the workspace, VSL is capable of extracting the ordered sequence of actions and reproducing the same task as expected. As shown in Figure~\ref{fig:reproVSL1}, the robot achieves the goal of the task using VSL. The accompanying video shows the execution of this task~\citep{icra2015video}.

\subsubsection*{Keeping All Objects Near}
\label{subsubsec:taskSecondExp}

\begin{figure}[H]
    \centering
    \includegraphics[trim=0cm 0cm 0cm 0cm, width=\columnwidth]{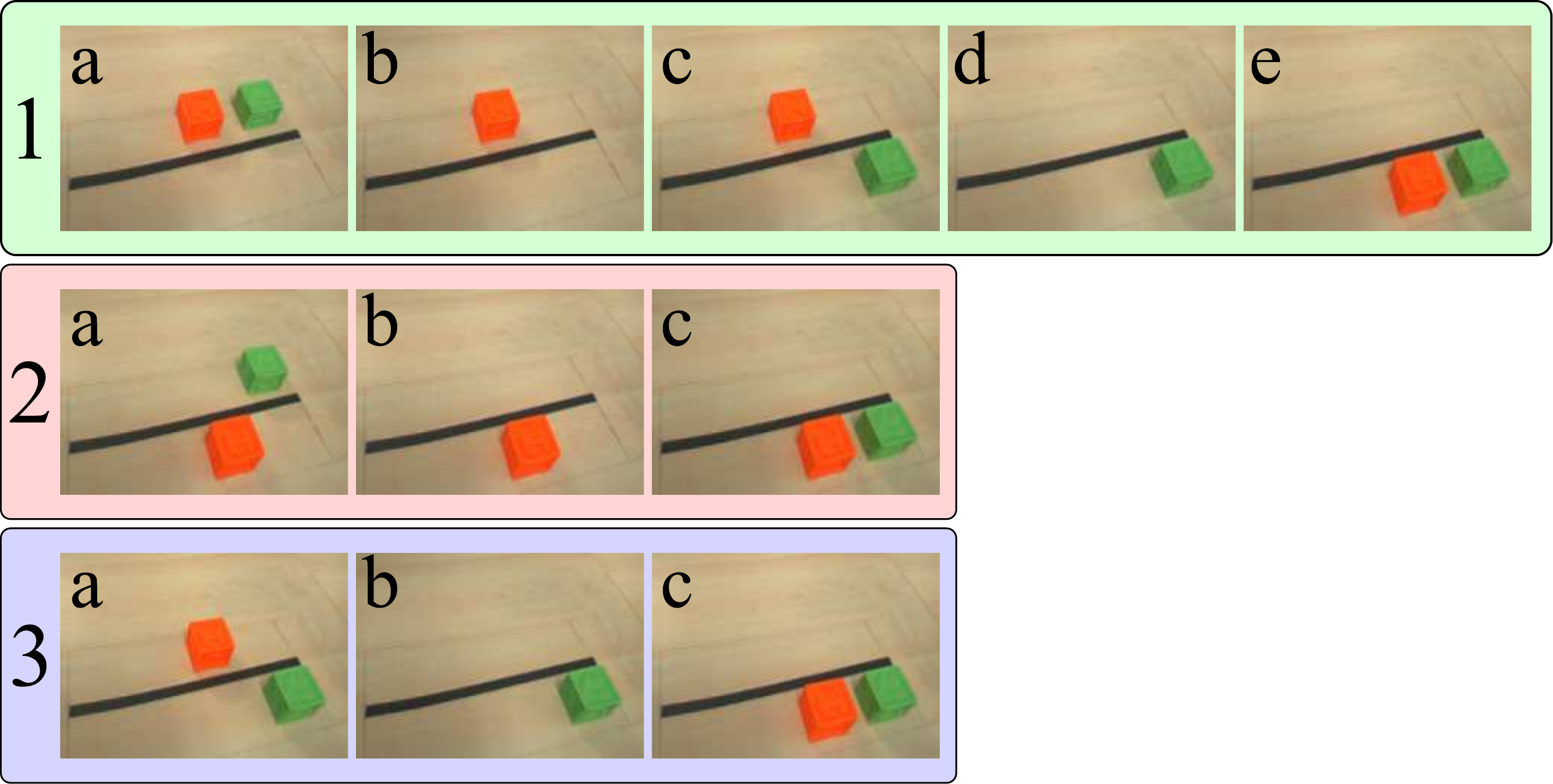}%
    \caption{Three different sets of demonstrations devised by the tutor for implying the rule of the game in the second experiment.}
    \label{fig:demoKeepAllNear}%
\end{figure}

In the previous experiment, it has been shown that the internal planner of VSL is capable of reproducing the task. However, VSL-SP is more capable in generalizing the learned skill. In this experiment, the robot learns a simple game including an implicit rule: \textit{keep all objects near}. For each object, two properties are defined: color and position. The symbolic planner provides VSL-SP with the capability of generalizing over the number and even shape of the objects. In order to utilize this capability, the tutor performs a set of demonstrations to eliminate the effect of color implying that any object independent of its color should not be \texttt{far} from the robot. Three sets of demonstrations are performed by the tutor, which are shown in Figure~\ref{fig:demoKeepAllNear}. Each demonstration starts with a different initial configuration of objects. The performed demonstrations imply the following rules:

\begin{equation}%
\begin{aligned}
&r_1^1: \langle B_{green}\rightarrow \texttt{near}\rangle \; \mathrm{IF} \; \langle B_{orange} \; is \; \texttt{far} \rangle \\
&r_1^2: \langle B_{orange}\rightarrow \texttt{near}\rangle \; \mathrm{IF} \; \langle B_{green} \; is \; \texttt{near} \rangle \\
&r_2^1: \langle B_{green}\rightarrow \texttt{near}\rangle \; \mathrm{IF} \; \langle B_{orange} \; is \; \texttt{near} \rangle \\
&r_3^1: \langle B_{orange}\rightarrow \texttt{near}\rangle \; \mathrm{IF} \; \langle B_{green} \; is \; \texttt{far} \rangle \\
\end{aligned}
\label{equ:rules1}%
\end{equation}

\noindent where $r_i^j$ indicates the $j^{\mathrm{th}}$ rule extracted from the $i^{\mathrm{th}}$ demonstration.
The first demonstration includes two rules $r_1^1,r_1^2$ because two actions are executed. It can be seen that the right parts of $r_1^1$ and $r_2^1$ and the right parts of  $r_1^2$ and $r_3^1$ eliminate each other. Also, the color attribute on the left side of the rules eliminates the effect of the color on action execution. During the demonstration, for each object, the robot extracts the spatial relationships between the objects and the borderline, and decides if the object is \texttt{far} or \texttt{near} (i.e. \texttt{not far}). It then applies the extracted precondition of $r_1^1: \langle B_{?}\rightarrow \texttt{near}\rangle$ to that object. Afterward, the robot is capable of performing the learned skill on a various number of objects and even unknown objects with different colors. The accompanying video shows the execution of this task~\citep{icra2015video}.

\subsubsection*{Pull All Objects in a Given Order}
\label{subsubsec:taskThirdExp}

In the previous experiment, the robot learns that all objects must be kept close, without giving importance to the order of actions. In this experiment, to give priority to the sequence of actions to be performed, the color attribute for each object is included as well. For instance, consider the ordered set $\langle orange, green, yellow\rangle$ in which the orange cube and the yellow cube are the most and least important cubes, respectively. This means that we want the orange cube to be moved before the green one, and the green cube before the yellow one. The robot has to learn that the most important cube has to be operated first.
It has been shown that learning a sequence of actions in which the order is important, is one of the main capabilities of VSL. Therefore, if the task was demonstrated once and the reproduction part was done by VSL (and not by VSL-SP) the robot would learn the task using one single demonstration. In addition, VSL provides the robot with the capability of generalizing the task over the initial objects' configuration.

Furthermore, using VSL-SP, it is possible to encode such an ordering also in a planning domain, at the cost of adding extra constraints (in the form of predicates) to the definition of the planning problem. In this case the planner should include two extra predicates, namely \texttt{(somewhere-after (far $B_{green}$) (far $B_{orange}$))} and \texttt{(somewhere-after (far $B_{yellow}$) (far $B_{green}$))}. The inclusion of these predicates must be managed at an extra-logic level, e.g. by extracting the sequence of actions using VSL and inserting the corresponding constraints in the planner problem definition. In this case, there is no need to have more demonstrations, but expressing the order of sequence as preconditions may be done automatically. The accompanying video shows the execution of this task~\citep{icra2015video}.

\begin{figure}[H]
    \centering
    \includegraphics[trim=0cm 0cm 0cm 0cm, width=\columnwidth]{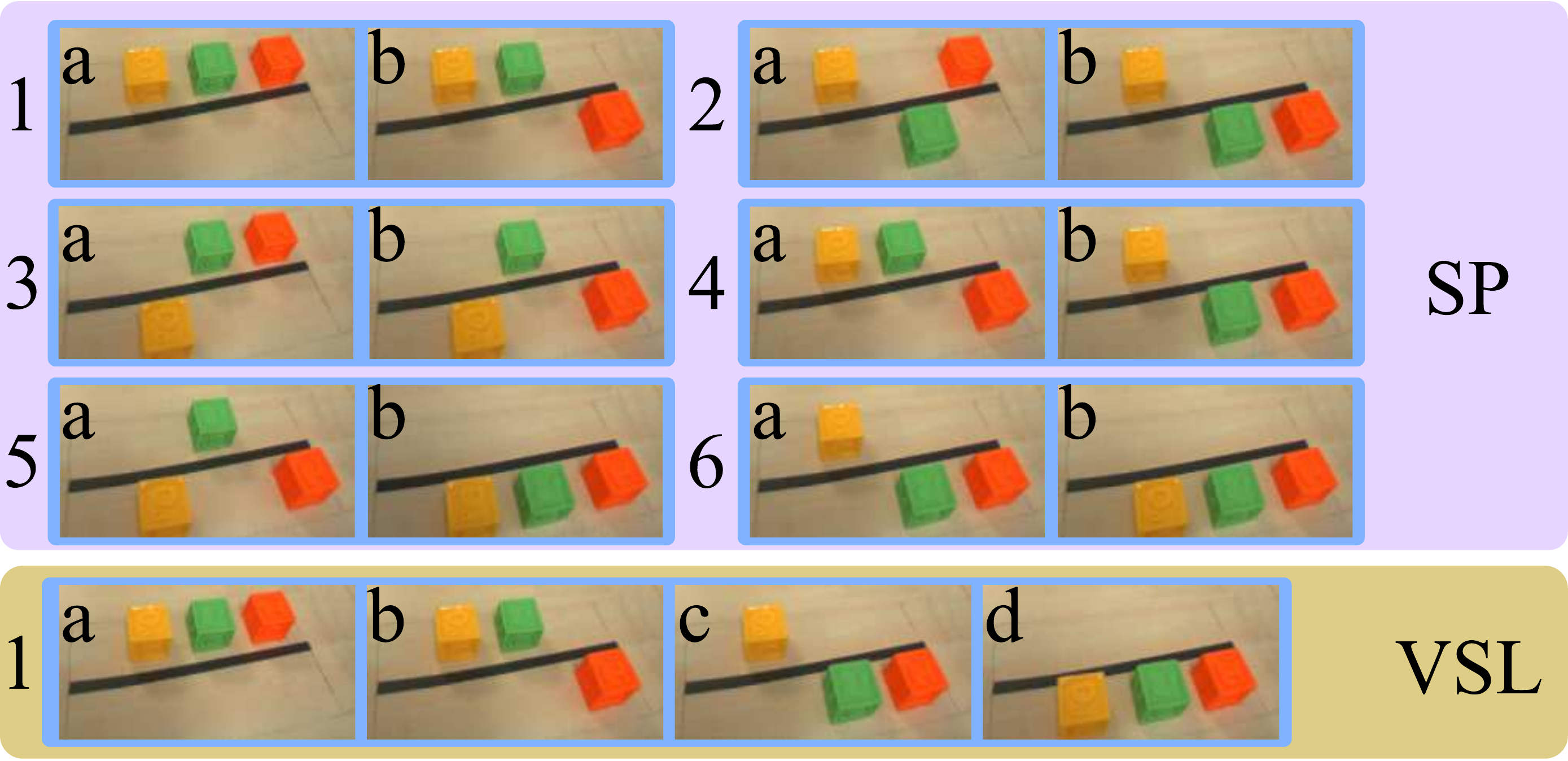}%
    \caption{In the third experiment, using VSL-SP, the robot needs six sets of demonstrations for each of which the initial (a) and final (b) objects' configurations are shown. Using VSL only the robot requires a single demonstration for which the sequence of operations is shown (a-d).}
    \label{fig:demoExpOrder}%
\end{figure}

\section{Conclusions}
\label{sec:conclusions}

In this paper, a novel skill learning approach has been proposed, which allows a robot to acquire new skills for object manipulation by observing a demonstration. VSL utilizes visual perception as the main source of information. As opposed to conventional trajectory-based learning approaches, VSL is a goal-based robot learning approach that focuses on achieving the desired goal configuration of objects relative to one another while maintaining the sequence of operations. VSL is capable of learning and generalizing multi-operation skills from a single demonstration while requiring minimum a priori knowledge about the environment.
It has been shown that integrating VSL with a conventional trajectory-based approach eliminates the need to program the primitive actions manually and to tune them according to the desired task. The robot learns the required primitive actions through kinesthetic teaching using Imitation Learning. Then it learns the sequence of actions to reach the desired goal of the task through VSL. The obtained framework easily extends and adapts robot’s capabilities to novel situations, even by users without programming ability.
It has been illustrated that to utilize the advantages of standard symbolic planners, VSL's original planner can be replaced with an already available action-level symbolic planner. In order to ground the actions, the symbolic actions are defined in the planner and VSL maps identified preconditions and effects in a formalism suitable to be used at the symbolic level. Experimental results have shown the improvement in generalization to find a solution that can be acquired from both multiple and single demonstrations.
In contrast to many existing approaches, VSL leverages simplicity, efficiency, and user-friendly human-robot interaction. The feasibility and efficiency of VSL have been validated through simulated and real-world experiments.

\section*{References}

\bibliography{Rezabib2}

\end{document}